%% file: main.tex
\pdfoutput=1

\documentclass[11pt]{article}

\PassOptionsToPackage{dvipsnames}{xcolor}
\usepackage{acl}

\usepackage{times}
\usepackage{latexsym}

\usepackage[T1]{fontenc}
\usepackage[utf8]{inputenc}

\usepackage{microtype}
\usepackage{inconsolata}
\usepackage{graphicx}
\usepackage{booktabs}
\usepackage{tabularx}
\usepackage{array}
\usepackage{subcaption}
\usepackage{float}
\usepackage{amsmath}
\usepackage{amssymb}
\usepackage{amsfonts}
\usepackage{multirow}
\usepackage[ruled,vlined]{algorithm2e}
\usepackage{xcolor}   
\usepackage{colortbl}

\usepackage{tcolorbox}
\tcbuselibrary{breakable}

\definecolor{blue}{RGB}{70,130,180}
\definecolor{green}{RGB}{34,139,34}
\newtcolorbox{promptBox}[4][]{
    colback=#4!10,
    colframe=#4!50!black,
    title=#2 #3,
    fonttitle=\bfseries,
    left=2mm,
    right=2mm,
    top=2mm,
    bottom=2mm,
    breakable,
    #1
}
\newcommand{\mytab}{\hspace{1em}}

\newcommand{\shortname}{\textbf{\textsc{CompassLLM}}}
\title{\shortname{}: A Multi-Agent Approach toward Geo-Spatial Reasoning for Popular Path Query}

\author
    {Md. Nazmul Islam Ananto$^{1}$ \quad Shamit Fatin$^{1,2}$ \quad Mohammed Eunus Ali$^{3}$ \\
    {\bf Md Rizwan Parvez}$^{4}$ \\
        $^{1}$Bangladesh University of Engineering and Technology (BUET) \\
        $^{2}$University of Utah \quad $^{3}$Monash University \quad $^{4}$Qatar Computing Research Institute (QCRI) \\
        nazmulislamananto@gmail.com , shamit.f@utah.edu ,\\
        eunus.ali@monash.edu , mparvez@hbku.edu.qa
    }

\begin{document}
\maketitle
\begin{abstract}
\input{sections/abstract}

\end{abstract}
\section{Introduction}
\input{sections/introduction}

\section{Related Works}\label{sec:related_works}
\input{sections/relatedworks}

\section{Problem Definition}\label{sec:problem_def}
\input{sections/problem_definition}

\section{Methodology}\label{sec:methodology}
\input{sections/method_agent}
\section{Experiments}\label{sec:experiment}
\input{sections/experimentalsettings}
\section{Results}\label{sec:result}
\input{sections/resultanalysis}
\section{Conclusion \& Future Works}\label{sec:conclusion}
\input{sections/conclusion}
\section*{Limitations}\label{sec:limitation}
\input{sections/limitations}

\bibliography{anthology,custom,custom_2}

\appendix
\input{sections/z_appendix}

\end{document}

%% file: sections/abstract.tex
The popular path query—identifying the most frequented routes between locations from historical trajectory data—has important applications in urban planning, navigation optimization, and travel recommendations. While traditional algorithms and machine learning approaches have achieved success in this domain, they typically require model training, parameter tuning, and retraining when accommodating data updates. As Large Language Models (LLMs) demonstrate increasing capabilities in spatial and graph-based reasoning, there is growing interest in exploring how these models can be applied to geo-spatial problems.

We introduce \shortname{}, a novel multi-agent framework that intelligently leverages the reasoning capabilities of LLMs into the geo-spatial domain to solve the popular path query. \shortname{} employs its agents in a two-stage pipeline: the \textit{SEARCH} stage that identifies popular paths, and a \textit{GENERATE} stage that synthesizes novel paths in the absence of an existing one, a common problem in sparse historical data. Experiments on real and synthetic datasets show that \shortname{} demonstrates superior accuracy in \textit{SEARCH} and competitive performance in \textit{GENERATE} both while being cost-effective.

%% file: sections/introduction.tex
\input{figures/exampleGeneration}

Recent advancements in Large Language Models (LLMs) have demonstrated remarkable versatility, excelling not only in natural language tasks but increasingly in complex reasoning problems across diverse domains \cite{laskar2024systematic}. Their zero-shot and in-context learning capabilities enable adaptation to new problem spaces without system modifications, making them attractive candidates for exploring challenging computational problems. As LLMs continue to show promise in different geo-spatial reasoning tasks and map data understanding \cite{dihan2025mapeval, hasan2025mapagenthierarchicalagentgeospatial, chai2023graphllmboostinggraphreasoning, roberts2023gpt4geolanguagemodelsees, balsebre2024lamplanguagemodelmap, xie2024travelplannerbenchmarkrealworldplanning, deng2025can}, there is growing interest in understanding how these capabilities can be harnessed for navigation and route recommendation tasks.

One compelling route recommendation task is the popular path query problem—mapping the most frequented routes between key locations (i.e. source-destination pair) from historical trajectory data (i.e. previously traveled paths by people). This problem
has significant applications in urban planning, navigation optimization, robotics, autonomous vehicle operations, and personalized (and non personalized) travel recommendations \cite{zheng2011urban, li2010swarm, yuan2011driving, yuan2010t, Giannotti2007TrajectoryPM}.

The problem becomes particularly interesting when considering path synthesis scenarios -- cases where no direct path exists in the historical trajectory data. In sparse trajectory dataset
like Figure \ref{fig:exampleGeneration}, where only 25\% of all possible edges exist (only $9$ edges out of $9C2=36$ possible edges for $9$ nodes)
in the dataset, generating novel path that is valid and traversable becomes a challenge. Generating valid paths is specifically crucial in this era of autonomous vehicles and other physical systems where LLMs' reasoning are being adopted \cite{latif20243pllmprobabilisticpathplanning, xiao2025llmadvisorllmbenchmarkcostefficient, doma2024llmenhancedpathplanningsafe}.

Traditional algorithmic approaches, including heuristic and approximation methods \cite{chen2011discovering, luo2013finding, wei2012constructing}, have achieved reasonable success in this domain. Machine learning approaches have also made progress, with earlier methods focusing on existing paths in historical data \cite{rashid2023deepalttrip, yang2018fast, wang2019empowering, tian2023effective, shi2024graphconstrained}, and more recent neural network models like NeuroMLR \cite{jain2021neuromlr} and GEIT \cite{ZHANG2023103176} capable of generating paths on graph. While these approaches are effective, they typically require model training, parameter tuning, and retraining when accommodating new constraints or data updates. Due to the lack of generalization in these approaches, LLM-based methods are getting increasing attention for their adaptability. LLMs' ability to process complex, long-context inputs and reason about relationships in general makes them intriguing candidates for spatial problems.

However, applying LLMs to geo-spatial reasoning presents unique challenges. While Chain-of-Thought (CoT) prompting \cite{cot} and its extensions like Self-Consistency \cite{wang2023selfconsistency}, ReAct \cite{yao2023react}, and Reflexion \cite{shinn2023reflexionlanguageagentsverbal} have shown success in various reasoning tasks, they are not crafted for geo-spatial tasks—particularly the popular path query—where LLMs may generate invalid or untraversable paths \cite{aghzal2023can} without proper guidance and direction. Recent efforts like LLM-A* \cite{meng2024llm} have combined algorithmic approaches with LLMs but focus on general pathfinding rather than popular path queries. Later, multiple researches including \cite{li2024research} and LLMAP \cite{yuan2025llmapllmassistedmultiobjectiveroute} utilizes LLM to parse user requirements and other parameters from the input. Most recently, PathGPT \cite{marcelyn2025pathgpt} explores path recommendation with retrival augmentation to embed only relevant part of the dataset with prompts.

To explore LLMs' potential in this domain, we introduce \shortname{} (Coordinated Orchestration of Multi-agent Path Analysis and Spatial Synthesis), a multi-agent framework designed specifically for popular path queries. Our approach features a novel two-stage pipeline orchestrating our agents: a \textit{SEARCH} stage that finds popular paths from historical trajectories, and a \textit{GENERATE} stage that synthesizes new paths when direct routes don't exist. \shortname{} leverages LLMs' understanding and reasoning capabilities of maps/graph structures.

Our evaluation uses both real-world trajectory data from user movement tracking \cite{36046891aa77462298f21dacee3279a6,10.1007/s10115-017-1056-y} and carefully designed synthetic datasets that capture diverse spatial configurations. Results demonstrate that \shortname{} outperforms other methods in \textit{SEARCH} problem and achieves competitive performance for the \textit{GENERATE} problem with SOTA models. All while offering the advantages of training-free inference and cost-effective token usage—particularly excelling in scenarios with sparse historical data.

Our paper makes the following contributions:

\begin{itemize}
    \item We propose \textit{\shortname{}}, a novel multi-agent framework that explores how LLMs can be effectively applied to geo-spatial domain for popular path queries.

    \item We incorporate path finding and synthesis in an orchestrated two-stage pipeline to recommend both popular and valid paths even in the cases of no historical ground truth.

    \item We provide comprehensive experimental evaluation demonstrating \shortname{}'s competitive performance against existing methods across real-world and synthetic datasets, while highlighting the unique advantages of LLM-based approaches for spatial reasoning tasks.
\end{itemize}

%% file: figures/exampleGeneration.tex
\begin{figure}[t]
    \centering
    \resizebox{\linewidth}{!}{%
    \includegraphics[width=\textwidth]{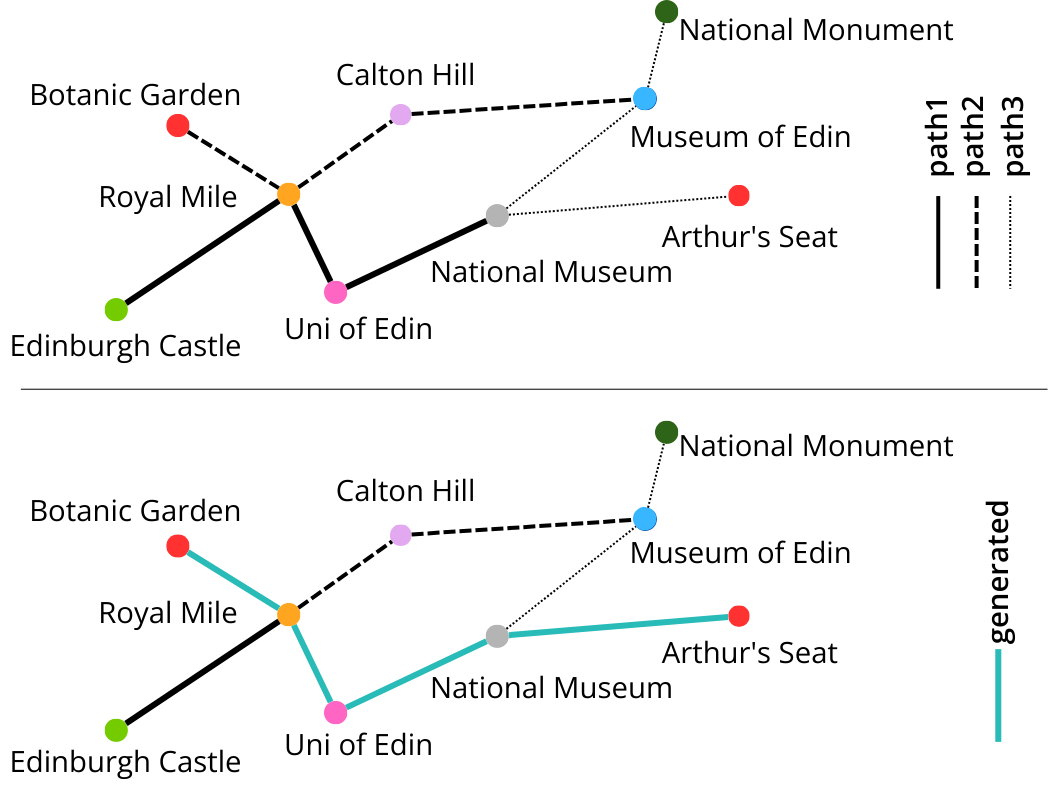}
    }
    \caption{
    When \textit{SEARCH}ing for the most popular path from \textbf{Botanic Garden} to \textbf{Arthur's Seat}, given the three trajectories (top), we can see that there is no single path spanning the entirety of these two. This incurs a \textit{GENERATE} problem (bottom) where one must break down the trajectories into smaller segments and combine them to form a new path.
    }
    \label{fig:exampleGeneration}
\end{figure}

%% file: sections/relatedworks.tex
\input{figures/overview}

The Popular Path Query in the form of Next POI Prediction or Route Recommendation has been a central research topic, driven by the increasing availability of trajectory data and the growing demand for personalized navigation services. Existing approaches can be broadly classified into three categories: algorithmic methods, machine learning models, and, more recently, LLM-based approaches.

\subsection*{Algorithmic Approaches}
Early research in this field predominantly employed probabilistic and heuristic methods \cite{banerjee2014inferring,chen2011discovering,luo2013finding,wei2012constructing,wu2016probabilistic}. However, these methods often require significant manual effort to handle corner cases and when new constraints are introduced.

\subsection*{Machine Learning Approach}
In recent years, machine learning has become the primary focus in route recommendation techniques \cite{wu2017modeling,li2020spatial,zheng2012reducing}. As they initially lacked path generation capabilities, deep generative models \cite{jain2021neuromlr,wu2017modeling} have since addressed this limitation. However, machine learning models remain inherently dependent on the data they are trained on, requiring time-consuming retraining. In scenarios like route recommendation, where data may change frequently due to environmental, natural, or economic factors, retraining becomes a significant bottleneck.

\subsection*{LLM-based Approaches}
Large Language Models have emerged as a revolutionary tool -- techniques such as prompt engineering \cite{wei2023chainofthought,wang2023selfconsistency,yao2023tree,yao2023react,shinn2023reflexionlanguageagentsverbal}, fine-tuning \cite{himoro-pareja-lora-2022-preliminary}, and prompt-tuning \cite{zhou2022large} have opened new avenues for LLMs to be applied in diverse fields. In particular, LLMs have demonstrated potential in areas such as graph reasoning \cite{chen2024exploring,ye2023natural,wang2024can} and task planning \cite{song2023llm,gundawar2024robust,sharan2023llm}. Nevertheless, challenges remain in ensuring that LLM-generated routes conform to real-world road networks \cite{kambhampati2024llms}. Key issues include the inherent inconsistency of LLM outputs \cite{jang-etal-2022-becel} and difficulties in parsing complex input structures \cite{tam2024let}. Despite these obstacles, the potential for LLMs to enhance the quality of route recommendations \cite{meng2024llm} and incorporate user satisfaction is considerable, indicating a promising direction for research \cite{li2024research, yuan2025llmapllmassistedmultiobjectiveroute, marcelyn2025pathgpt}.

%% file: figures/overview.tex
\begin{figure*}[ht]
    \centering
    \includegraphics[width=0.9\textwidth]{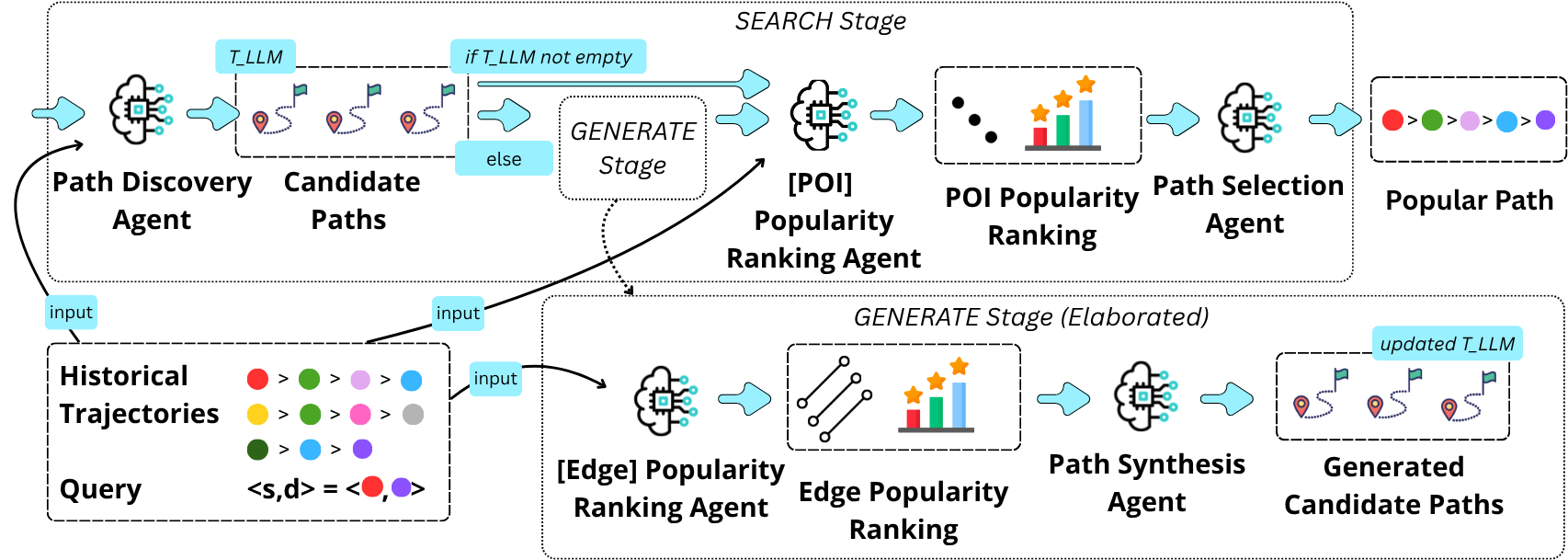}
    \caption{Overview of \shortname{} framework.}
    \label{fig:MainBanner}
\end{figure*}

%% file: sections/problem_definition.tex
Let $\mathcal{V} = \{p_1, p_2, \ldots, p_N \}$ be a set of $N$ POIs in a city (e.g., Edinburgh Castle, Calton Hill, National Monument). The historical trajectories, denoted as $\mathcal{T} = \{r_1, r_2, \ldots, r_M \}$, represent a collection of $M$ routes that people have taken while visiting the city in the past. Each route $r \in \mathcal{T}$ is an ordered sequence of POIs i.e. $r = (p_{1}, p_{2}, \ldots, p_{J})$, where each $(p_j,p_{j+1})$ is a road segment between two POIs. Moreover, $\mathcal{E_T} = \{(p_{j}, p_{j+1}): p_{j}, p_{j+1} \in r; \forall r \in \mathcal{T} \}$ denotes the set of all road segments present in $\mathcal{T}$.
As the true map is not available in the dataset, we construct the graph $ \mathcal{G} = (\mathcal{V}, \mathcal{E_T})$ from only the historical trajectories $\mathcal{T}$.

Given these historical trajectories $\mathcal{T}$ and a query $q:\langle s, d \rangle$ as input, our aim is to find the single most popular path $\mathcal{R} = (q_1, q_2, \ldots, q_K)$. This path should satisfy the conditions that $q_1 = s$, $q_K = d$ and $\forall k: q_k \in \mathcal{V}$.

The ground truth trajectories $\mathcal{T}_{GT} = \{r_1\prime, r_2\prime, \ldots\}$ for a given query $ q:\langle s, d \rangle $ is the list of unique paths or sub-paths, $r\prime$, extracted from each path $r \in \mathcal{T}$ such that $r_\prime = (s, \ldots, d)$.
The popularity of each path $r^\prime$ is defined as the cumulative popularity of its POIs:
$\mathrm{Popularity}(r^\prime) = \sum_{k=1}^{K} pop(q_k)$

Thus, the most popular path—defined here as the path maximizing cumulative POI popularity among candidates in $\mathcal{T}_{GT}(s,d)$—is obtained by:
\[
\mathcal{R} = \arg\max_{r^\prime \in \mathcal{T_{GT}}(s,d)} \sum_{q_k \in r^\prime} pop(q_k)
\]
This POI-additive formulation is distinct from exact route-frequency maximization as our objective ranks candidate paths ($T_{LLM}$ in Figure \ref{fig:MainBanner}, elaborated in Section \ref{sec:methodology}) by the cumulative popularity of their constituent POIs, following prior literature \cite{rashid2023deepalttrip}.

On the other hand, if $ \mathcal{T}_{GT} = \emptyset $, there is no path satisfying the given query, let alone a popular one. In such cases, our goal is to generate a popular path which is traversable in the graph $\mathcal{G}$. We consider a route $\mathcal{R}$ to be fully traversable if its constituent edges belong to $\mathcal{E_T}$.

%% file: sections/method_agent.tex
\input{tables/f1_comparison}

\input{tables/traversability_comparison}

\input{tables/cost_comparison}

We propose a multi-agent framework \cite{islam2024mapcoder, hong2023metagpt} for popular path queries that orchestrates specialized agents across a two-stage pipeline: \textit{SEARCH} and \textit{GENERATE}. Our approach coordinates four key agents---\textit{Path Discovery}, \textit{Popularity Ranking}, \textit{Path Synthesis}, and \textit{Path Selection}---each designed to handle distinct aspects of the popular path finding process.

Figure~\ref{fig:MainBanner} shows an overview of the framework. The orchestration starts with the \textit{SEARCH} stage. The \textit{Path Discovery Agent} first retrieves candidate paths from historical trajectories that satisfy the query $\langle s, d \rangle$. When candidate paths are available, the framework remains in the \textit{SEARCH} stage and proceeds directly to path evaluation. However, when no suitable paths exist in the historical data, the framework enters the \textit{GENERATE} stage.

In the \textit{GENERATE} stage, the \textit{Popularity Ranking Agent} operates in edge mode, computing popularity scores for individual edges based on historical traversal patterns. These rankings guide the \textit{Path Synthesis Agent} in constructing new candidate paths that connect the source and destination, and these candidates are then returned to the \textit{SEARCH} stage. Thus, the \textit{GENERATE} stage is conditionally executed within the \textit{SEARCH} stage, and our orchestration begins and ends with \textit{SEARCH}.

Upon returning to (or remaining in) the \textit{SEARCH} stage with candidate paths---whether discovered or generated---the \textit{Popularity Ranking Agent} is invoked in POI mode to rank locations by their popularity. Finally, the \textit{Path Selection Agent} evaluates all candidates using these POI rankings and returns the top-ranked path as the \textit{Popular Path}.

This orchestration enhances the overall spatial reasoning capability of the framework by leveraging each agent's specialization while maintaining flexibility through conditional stage transitions. In the following subsections, we discuss the agents, their responsibilities, and interactions.

\subsection{Path Discovery Agent}

The \textit{Path Discovery Agent} serves as the entry point, utilizes LLM to identify trajectory segments where the source appears before the destination, extracting candidate paths $\mathcal{T}_{LLM}$ without heavy computations or external retrieval models. We instruct the LLM to first identify occurrences of source $s$ and destination $d$ POIs within historical data $\mathcal{T}$, then extract complete path segments where $s$ precedes $d$. In cases of loops, the agent considers both the shorter and the longer segments. The agent either provides \textit{Candidate Paths} for evaluation (remaining in \textit{SEARCH} stage) or returns an empty set $\mathcal{T}_{LLM} = \emptyset$ that triggers the \textit{GENERATE} stage.

\subsection{Popularity Ranking Agent}

The \textit{Popularity Ranking Agent} quantifies spatial entity importance through frequency analysis, operating in two distinct modes: \textit{Edge ranking mode} and \textit{POI ranking mode}. In \textit{Edge ranking mode}, activated conditionally within the \textit{GENERATE} stage when no historical paths exist, we prompt the LLM to identify and rank edges from historical trajectories in descending order of occurrence frequency. These edge rankings guide the \textit{Path Synthesis Agent} in constructing novel paths. In \textit{POI ranking mode}, which is always executed within the \textit{SEARCH} stage---either directly after path discovery or following the \textit{GENERATE} stage execution---we prompt the LLM to rank all unique POIs in descending order of occurrence frequency.

\subsection{Path Synthesis Agent}

The \textit{Path Synthesis Agent} generates novel traversable paths when discovery fails, operating exclusively within the \textit{GENERATE} stage. This agent receives edge rankings from the \textit{Popularity Ranking Agent} and constructs valid paths by connecting popular edges, analogous to humans planning routes through unfamiliar areas by combining known road segments.
The prompt emphasizes using only edges present in edge popularity ranking; this encodes validity in the \textit{GENERATE} stage through an observed-edge constraint, eliminating hallucination of non-existent connections---a critical requirement for safety-critical navigation systems.
The resulting \textit{Generated Candidate Paths} update $\mathcal{T}_{LLM}$ and are returned to the \textit{SEARCH} stage for evaluation.

\subsection{Path Selection Agent}

The \textit{Path Selection Agent} performs the final path evaluation within the \textit{SEARCH} stage. It receives candidate paths $\mathcal{T}_{LLM}$ (either discovered or synthesized) along with POI rankings, then prompts the LLM to score each path by aggregating the popularity of its constituent POIs. The agent selects the highest-scoring path $\mathcal{R}$ as the final \textit{Popular Path}.

Appendix \ref{app:Algorithm} contains the algorithm of the entire pipeline and Appendix~\ref{app:prompts} provides the detailed prompts for each agent.

%% file: tables/f1_comparison.tex
\begin{table*}[!tb]
\centering
\resizebox{0.9\linewidth}{!}{%

\begin{tabular}{c|l|cccccccc}
\toprule
\multirow{2}{*}{\textbf{Type}} & \multirow{2}{*}{\textbf{Method}} & \textbf{DisHolly} & \textbf{Epcot} & \textbf{CaliAdv} & \textbf{Edin} & \textbf{Toro} & \textbf{Disland} & \textbf{Melb} & \textbf{Average} \\
& & 13 POIs & 17 POIs & 25 POIs & 28 POIs & 29 POIs & 31 POIs & 88 POIs & - \\
\midrule

\multirow{5}{*}{\rotatebox{90}{\shortstack{\textbf{ML/DL}\\\textbf{based}}}}
& Markov & 0.61 & 0.59 & 0.53 & 0.59 & 0.60 & 0.50 & 0.51 & 0.56 \\
& NASR & 0.63 & 0.60 & 0.54 & 0.58 & 0.58 & 0.50 & 0.53 & 0.57 \\
& DeepAltTrip-LSTM & 0.62 & 0.59 & 0.53 & 0.58 & 0.59 & 0.52 & 0.52 & 0.56 \\
& DeepAltTrip-Samp & 0.62 & 0.58 & 0.53 & 0.58 & 0.59 & 0.51 & 0.51 & 0.56 \\
& NMLR & 0.70 & 0.69 & 0.55 & 0.62 & 0.69 & 0.65 & 0.60 & 0.65 \\
\midrule

\multirow{8}{*}{\rotatebox{90}{\shortstack{\textbf{LLM based}\\\textbf{(Llama 3.1 8b)}}}}
& Direct & 0.50 & 0.46 & 0.47 & 0.45 & 0.45 & 0.43 & 0.40 & 0.45 \\
& CoT & 0.59 & 0.53 & 0.56 & 0.54 & 0.57 & 0.48 & 0.47 & 0.53 \\
& SC & 0.57 & 0.55 & 0.54 & 0.56 & 0.54 & 0.50 & 0.45 & 0.53 \\
& APE & 0.45 & 0.40 & 0.42 & 0.41 & 0.38 & 0.34 & 0.32 & 0.39 \\
& ReAct & 0.68 & 0.64 & 0.66 & 0.70 & 0.59 & 0.47 & 0.43 & 0.60 \\
& Reflexion & 0.69 & 0.64 & 0.66 & 0.70 & 0.59 & 0.48 & 0.44 & 0.60 \\
& LLM A* & 0.58 & 0.53 & 0.55 & 0.60 & 0.52 & 0.42 & 0.40 & 0.51 \\
& PathGPT & 0.68 & 0.61 & 0.57 & 0.74 & 0.76 & 0.57 & 0.63 & 0.65 \\
& \cellcolor{SkyBlue!50}\shortname{} & \cellcolor{SkyBlue!50}0.75 & \cellcolor{SkyBlue!50}0.69 & \cellcolor{SkyBlue!50}0.71 & \cellcolor{SkyBlue!50}0.76 & \cellcolor{SkyBlue!50}0.62 & \cellcolor{SkyBlue!50}0.49 & \cellcolor{SkyBlue!50}0.44 & \cellcolor{SkyBlue!50}0.64 \\
\midrule

\multirow{8}{*}{\rotatebox{90}{\shortstack{\textbf{LLM based}\\\textbf{(GPT 4o)}}}}
& Direct & 0.56 & 0.51 & 0.52 & 0.50 & 0.50 & 0.48 & 0.45 & 0.50 \\
& CoT & 0.66 & 0.57 & 0.61 & 0.59 & 0.62 & 0.51 & 0.52 & 0.58 \\
& SC & 0.64 & 0.59 & 0.58 & 0.61 & 0.59 & 0.53 & 0.49 & 0.58 \\
& APE & 0.79 & 0.71 & \textbf{0.75} & \underline{0.81} & 0.65 & 0.50 & 0.45 & 0.67 \\
& ReAct & 0.74 & 0.70 & 0.73 & 0.74 & 0.64 & 0.49 & 0.45 & 0.64 \\
& Reflexion & 0.74 & 0.70 & 0.73 & 0.74 & 0.64 & 0.51 & 0.46 & 0.65 \\
& LLM A* & 0.61 & 0.56 & 0.57 & 0.62 & 0.54 & 0.45 & 0.42 & 0.54 \\
& PathGPT & 0.54 & 0.43 & 0.34 & 0.54 & 0.58 & 0.24 & 0.44 & 0.44 \\
& \cellcolor{SkyBlue!50}\shortname{} & \cellcolor{SkyBlue!50}0.80 & \cellcolor{SkyBlue!50}0.73 & \cellcolor{SkyBlue!50}\underline{0.74} & \cellcolor{SkyBlue!50}\underline{0.81} & \cellcolor{SkyBlue!50}0.65 & \cellcolor{SkyBlue!50}0.50 & \cellcolor{SkyBlue!50}0.45 & \cellcolor{SkyBlue!50}0.67 \\
\midrule

\multirow{8}{*}{\rotatebox{90}{\shortstack{\textbf{LLM based}\\\textbf{(GPT o3 mini)}}}}
& Direct & 0.74 & 0.73 & 0.59 & 0.57 & 0.68 & 0.67 & 0.65 & 0.66 \\
& CoT & \underline{0.82} & 0.73 & 0.71 & 0.75 & 0.84 & 0.67 & 0.78 & 0.76 \\
& SC & 0.74 & \underline{0.76} & 0.57 & 0.68 & 0.84 & 0.64 & 0.80 & 0.72 \\
& APE & 0.81 & 0.74 & 0.71 & 0.80 & \underline{0.89} & 0.64 & \underline{0.81} & \underline{0.77} \\
& ReAct & 0.75 & 0.73 & 0.61 & 0.72 & 0.87 & 0.63 & 0.77 & 0.73 \\
& Reflexion & 0.75 & 0.75 & 0.65 & 0.77 & 0.81 & 0.52 & 0.77 & 0.72 \\
& LLM A* & 0.75 & 0.73 & 0.72 & 0.67 & 0.87 & \textbf{0.72} & 0.77 & \underline{0.77} \\
& PathGPT & 0.74 & 0.61 & 0.69 & 0.68 & 0.76 & 0.47 & 0.59 & 0.65 \\
& \cellcolor{SkyBlue!50}\shortname{} & \cellcolor{SkyBlue!50}\textbf{0.84} & \cellcolor{SkyBlue!50}\textbf{0.79} & \cellcolor{SkyBlue!50} {0.73} & \cellcolor{SkyBlue!50}\textbf{0.82} & \cellcolor{SkyBlue!50}\textbf{0.90} & \cellcolor{SkyBlue!50}\underline{0.71} & \cellcolor{SkyBlue!50}\textbf{0.82} & \cellcolor{SkyBlue!50}\textbf{0.80} \\
\bottomrule
\end{tabular}

}
\caption{
$F1$ scores comparison across datasets. For each dataset, best scores are bolded and second bests are underlined.
}
\label{tab:f1_sorted_avg}
\end{table*}

%% file: tables/traversability_comparison.tex
\begin{table}[ht]
\centering
\resizebox{\linewidth}{!}{%

\begin{tabular}{c|l|cccc}
\toprule
\multirow{2}{*}{\textbf{Type}} & \multirow{2}{*}{\textbf{Method}} & \textbf{Edin} & \textbf{Toro} & \textbf{Melb} & \textbf{Avg} \\
& & 28 POIs & 29 POIs & 88 POIs & - \\
\midrule

\multirow{1}{*}{\textbf{DL}}
& NMLR & \underline{0.89} & \textbf{0.98} & \underline{0.95} & \underline{0.94} \\
\midrule

\multirow{8}{*}{\rotatebox{90}{\shortstack{\textbf{LLM based}\\\textbf{(Llama 3.1 8b)}}}}
& Direct & 0.12 & 0.22 & 0.17 & 0.17 \\
& CoT & 0.21 & 0.23 & 0.19 & 0.21 \\
& SC & 0.19 & 0.22 & 0.21 & 0.21 \\
& APE & 0.53 & 0.57 & 0.42 & 0.51 \\
& ReAct & 0.56 & 0.51 & 0.63 & 0.57 \\
& Reflexion & 0.71 & 0.77 & 0.78 & 0.75 \\
& LLM A* & 0.78 & 0.82 & 0.49 & 0.70 \\
& PathGPT & 0.60 & 0.53 & 0.17 & 0.43 \\
& \cellcolor{SkyBlue!50}\shortstack{\shortname{}} & \cellcolor{SkyBlue!50}0.72 & \cellcolor{SkyBlue!50}0.78 & \cellcolor{SkyBlue!50}0.61 & \cellcolor{SkyBlue!50}0.70 \\
\midrule

\multirow{8}{*}{\rotatebox{90}{\shortstack{\textbf{LLM based}\\\textbf{(GPT 4o)}}}}
& Direct & 0.30 & 0.21 & 0.33 & 0.28 \\
& CoT & 0.48 & 0.60 & 0.20 & 0.43 \\
& SC & 0.50 & 0.60 & 0.22 & 0.44 \\
& APE & 0.86 & 0.89 & 0.94 & 0.90 \\
& ReAct & 0.82 & 0.85 & 0.86 & 0.84 \\
& Reflexion & 0.83 & 0.86 & 0.88 & 0.86 \\
& LLM A* & 0.84 & 0.82 & 0.90 & 0.85 \\
& PathGPT & 0.83 & 0.78 & 0.43 & 0.68 \\
& \cellcolor{SkyBlue!50}\shortstack{\shortname{}} & \cellcolor{SkyBlue!50}0.84 & \cellcolor{SkyBlue!50}0.91 & \cellcolor{SkyBlue!50}\textbf{0.96} & \cellcolor{SkyBlue!50}0.90 \\
\midrule

\multirow{8}{*}{\rotatebox{90}{\shortstack{\textbf{LLM based}\\\textbf{(GPT o3 mini)}}}}
& Direct & 0.54 & 0.51 & 0.53 & 0.53 \\
& CoT & 0.56 & 0.51 & 0.56 & 0.54 \\
& SC & 0.83 & 0.56 & 0.53 & 0.64 \\
& APE & 0.80 & 0.71 & 0.88 & 0.80 \\
& ReAct & 0.69 & 0.79 & 0.81 & 0.76 \\
& Reflexion & 0.54 & 0.63 & 0.91 & 0.69 \\
& LLM A* & 0.56 & 0.53 & 0.90 & 0.66 \\
& PathGPT & \underline{0.89} & 0.86 & 0.71 & 0.82 \\
& \cellcolor{SkyBlue!50}\shortstack{\shortname{}} & \cellcolor{SkyBlue!50}\textbf{0.95} & \cellcolor{SkyBlue!50}\underline{0.94} & \cellcolor{SkyBlue!50}\underline{0.95} & \cellcolor{SkyBlue!50}\textbf{0.95} \\
\bottomrule
\end{tabular}

}
\caption{
$Traversability$ scores across datasets. For each dataset, best scores are bolded and second bests are underlined.
}
\label{tab:traversability}
\end{table}

%% file: tables/cost_comparison.tex
\begin{table*}[ht]
\centering
\resizebox{0.9\linewidth}{!}{%

\begin{tabular}{llcccccccc}
\toprule
\textbf{Metric} & \textbf{Method} & \textbf{DisHolly} & \textbf{Epcot} & \textbf{CaliAdv} & \textbf{Edin} & \textbf{Toro} & \textbf{Disland} & \textbf{Melb} & \textbf{Average} \\
\midrule
\multirow{4}{*}{Token Count\ \ }
& APE        & 5.4k  & 8.7k  & 11.6k & 17.7k & 25.5k & 32.5k & 67.7k & 24.2k \\
& LLM-A*     & \underline{2.8k} & \underline{4.3k} & \textbf{5.6k} & \textbf{8.5k} & \textbf{12.0k} & \textbf{15.3k} & \textbf{31.6k} & \textbf{11.4k} \\
& Reflexion  & 10.1k & 15.6k & 20.5k & 31.5k & 46.2k & 59.7k & 99.9k & 40.5k \\
\rowcolor{SkyBlue!50} & \shortname{}  & \textbf{2.6k} & \textbf{4.2k} & \underline{5.7k} & \underline{8.6k} & \underline{12.4k} & \underline{15.8k} & \underline{33.0k} & \underline{11.8k} \\
\bottomrule
\end{tabular}

}
\caption{Cost comparison among different LLM-based approaches. Best scores are bolded and second bests are underlined.}
\label{tab:prompt_comparison}
\end{table*}

%% file: sections/experimentalsettings.tex

\subsection{Dataset}

\paragraph{Real-world Data:}
We used seven Real-world datasets \cite{rashid2023deepalttrip} across two domains: geo-tagged Flickr traces from Edinburgh (\textbf{Edin}), Toronto (\textbf{Toro}), and Melbourne (\textbf{Melb}) \cite{36046891aa77462298f21dacee3279a6}, and trip data from four theme parks, Disney's Hollywood Studios (\textbf{DisHolly}), Epcot Theme Park (\textbf{Epcot}), Disney California Adventure (\textbf{CaliAdv}) \& Disneyland (\textbf{Disland}) \cite{10.1007/s10115-017-1056-y}.
While all the datasets were used for \textit{SEARCH} problem, only \textbf{Edin}, \textbf{Toro} \& \textbf{Melb} were used for the \textit{GENERATE} problem as other datasets didn't have any POI pair that doesn't have any ground truth, thus unsuitable for \textit{GENERATE} problem evaluation.

\label{para:synthetic_data}
\paragraph{Synthetic Data:}
We generated synthetic datasets to introduce spatial diversity, addressing the edge count and edge distribution in real data. Using the reverse-delete algorithm, we controlled edge density while a Steiner tree approximation \cite{robins2000improved} emphasized key "highway" routes. Users followed these highways 90\% of the time, deviating only when shorter paths were available, closely mirroring real-world movement patterns.
Appendix~\ref{app:SyntheticData} provides the entire algorithm and validation for synthetic data generation in detail.

\subsection{Baselines \& Implementation}
We evaluated \shortname{} against traditional machine learning, deep learning, and LLM-based models. The \textbf{Markov} model \cite{chen2016learning} predicts user movements with Markov chains, while \textbf{NASR} \cite{wang2019empowering} employs neural architecture search for urban navigation. \textbf{DeepAltTrip} \cite{rashid2023deepalttrip} generates diverse alternative routes, and \textbf{NMLR} \cite{jain2021neuromlr} creates paths for large-scale traffic networks using Graph Neural Network. Among LLM techniques, we benchmarked \textbf{Direct} (Zero Shot), \textbf{CoT} (Chain of Thought) \cite{wei2023chainofthought} which breaks tasks into sequential steps, \textbf{SC} (Self-Consistency) \cite{wang2023selfconsistency} which selects the most consistent reasoning paths, and \textbf{APE} \cite{zhou2022large} which automates task-specific prompt generation. \textbf{ReAct} \cite{yao2023react} integrates reasoning with action, \textbf{Reflexion} \cite{shinn2023reflexionlanguageagentsverbal} enhances performance through iterative feedback, and \textbf{LLM A*} \cite{meng2024llmalargelanguagemodel} adapts A* search for improved spatial reasoning. Finally, \textbf{PathGPT} \cite{marcelyn2025pathgpt} implements RAG for retrieving relevant historical data to be sent with zero-shot prompting.

For each method, hyperparameters like \textit{temperature} were set and prompts were crafted following the respective guidelines of the technique. We recorded the number of tokens used in the prompt as well as the tokens generated in the response. For LLM-based methods, we used both an open source (Llama 3.1 8B) and two closed-source models (GPT 4o \& GPT o3 mini) as the underlying language model.
Additionally, a medium \textit{reasoning effort} was used for the reasoning model, GPT o3 mini.
For \shortname{}, we maintained a lower \textit{temperature} across all experiments and models as it shows better results [Appendix \ref{app:avg_temp}].

Classical graph-search algorithms would be NP-hard for our objective \cite{casel_fine-grained_2023, martens_complexity_2022}, as maximizing POI-additive popularity over simple paths between fixed $\langle s, d \rangle$ reduces to a maximum-weight simple path problem. Similarly, popularity-weighted graph-search variants rely on heuristic design choices (edge weights, normalization, length-bias control) rather than serving as exact solvers for the popular-path objective. For these reasons, we do not include them as baselines.

\subsection{Evaluation Metrics}\label{subsec:metric}
We used two key metrics: $F1$ for \textit{SEARCH} problems \cite{rashid2023deepalttrip} and $Traversability$ for \textit{GENERATE} problems.

The $F1$, a harmonic mean of precision and recall, evaluates the accuracy of the recommended paths by comparing them to the ground truth popular paths in the historical trajectories in $\mathcal{T}$.

Let $\mathcal{R} = (q_1, q_2,\ldots, q_K)$ be a recommended path. Also, let $\mathcal{V}_r$ and $\mathcal{V}_{\mathcal{R}}$ denote the sets of POIs in the ground-truth popular path and the recommended path, respectively. The precision($P$), recall ($R$), and $F1$ of the recommended itinerary $\mathcal{R}$ is calculated as:
\begin{align*}
    P = \frac{|\mathcal{V}_r \cap \mathcal{V}_{\mathcal{R}}|}{|\mathcal{V}_{\mathcal{R}}|};
    R = \frac{|\mathcal{V}_r \cap \mathcal{V}_{\mathcal{R}}|}{|\mathcal{V}_r|};
    F1 = \frac{2*P*R}{P + R}
\end{align*}
The $Traversability$ metric is an improved version of $Reachability$ from \textbf{NMLR} \cite{jain2021neuromlr}. While $Reachability$, being a binary metric, only checks if the last POI in the recommended path is the query destination or not, $Traversability$ measures the model's ability to generate paths that connect source and destination points using consecutive road segments that actually exist in historical routes. For a recommended route $\mathcal{R}=(q_1, q_2,
\ldots, q_K)$ and the set of all road segments $\mathcal{E_T}$, it is mathematically defined as,
\begin{align*}
T&raversability = \\
&\frac{|\{(q_k, q_{k+1}) : (q_k, q_{k+1}) \in \mathcal{E}_T, k \in [1, K-1]\}|}{K-1}
\end{align*}

A higher $Traversability$ score indicates a higher degree of edge validity, particularly crucial for safety-critical systems.

%% file: sections/resultanalysis.tex
\input{figures/synthetic_data_eval}

\subsection*{Performance Analysis}

According to established literature in this domain \cite{jain2021neuromlr, rashid2023deepalttrip}, models aim for even slight improvement in accuracy. In Table \ref{tab:f1_sorted_avg} we can see that LLM-based methods consistently outperform traditional ML/DL-based methods in the \textit{SEARCH} phase. \textbf{APE} performs well on GPT o3 mini \& GPT 4o but struggles with Llama 3.1 8b due to its smaller model size; as this method is feedback based, performance in each stage is accumulated. \textbf{LLM-A*} struggles with smaller model as well.
On the other hand, \textbf{ReAct} and \textbf{Reflexion} can perform consistently with smaller models too. But both \textbf{APE} and \textbf{LLM-A*} outperform these approaches with GPT o3 mini, an actual reasoning model. Notably, \shortname{} outperforms other methods in almost all model-specific implementations, achieving high-quality results with consistency. As GPT o3 mini is a reasoning model, it has an overall better performance over GPT 4o and Llama 3.1 8b.

In table \ref{tab:traversability}, only \textbf{NMLR} is used among ML/DL-based methods, as others failed to \textit{GENERATE} paths without ground truth. It is worth noticing that \textbf{NMLR} is specifically built to synthesize valid paths-thus score high on metrics like $Traversability$ with low score on $F1$. However, this comes at the cost of requiring significant retraining on new data, limiting their adaptability in dynamic environments. \textbf{PathGPT} performs well in small datasets (Edin) but struggle more than other methods as the dataset grows large (Melb).
In contrast, \shortname{} maintains competitive $Traversability$ performance with \textbf{NMLR} while outperforming others.

This establishes \shortname{} as an obvious choice for popular path queries offering (1) Significant practical benefits of LLMs, requiring fewer resources and retraining in dynamic datasets, (2) Outperforming others in \textit{SEARCH} and matching SOTA in \textit{GENERATE}, better of both worlds. \shortname{} is most beneficial under sparsity and missing-route scenarios; simpler methods may be competitive in dense regimes. Additional evaluations, including statistical significance and variance analysis, are presented in Appendix \ref{app:additional_results}.

\subsection*{Cost Analysis}

A key advantage of \shortname{} is its ability to balance performance with computational efficiency, as demonstrated in Table \ref{tab:prompt_comparison} (keeping only the best performing methods from Table \ref{tab:f1_sorted_avg}).
\textbf{APE}, while achieving strong results, incurs significantly higher computational costs due to its reliance on extensive prompt engineering. For instance, in dataset \textbf{Melb} with the largest number of POIs, \textbf{APE}'s token count reaches 67.7 thousand. Similarly, \textbf{Reflexion} also uses a lot of tokens and costs more than the other two. On the other hand, \textbf{LLM-A*} is the most cost-effective method but sometimes struggles with performance as seen in table \ref{tab:f1_sorted_avg} \& \ref{tab:traversability}, making it unsuitable for high-quality path generation. The efficiency of \shortname{} becomes even more pronounced in datasets with smaller POIs, where it incurs costs similar to \textbf{LLM-A*} while outperforming it in reasoning for popular path.

\subsection*{Scalability Studies}

We evaluate scalability using synthetic datasets with varying complexity: 10-50 POIs and 50-150 trajectories. Figure~\ref{fig:combined_figures} compares \shortname{} against the best-performing ML/DL baseline and competitive LLM-based methods from table \ref{tab:f1_sorted_avg}, \ref{tab:traversability} \& \ref{tab:prompt_comparison}. A general downward trend is observed as either of \#POI or \#trajectories increases. Higher \#POI implies sparse data and increased \#trajectories denote a larger dataset.

As \#POIs increases (top row), $F1$ scores decline across all methods, reflecting the increased difficulty of sparse networks. \shortname{} maintains superior $F1$ performance, outperforming \textbf{Best ML/DL} by \textbf{14\%} at 40 POIs and by \textbf{11.5\%} at 50 POIs. More notably, \shortname{} achieves \textbf{29\%} higher $Traversability$ than \textbf{APE} at 30 to 50 POIs, while \textbf{LLM-A*} degrades significantly in both metrics.

With increasing \#trajectories (bottom row), \shortname{} maintains a consistent \textbf{6-10\%} $F1$ advantage over \textbf{Best ML/DL} baselines across all settings. For $Traversability$, \shortname{} sustains \textbf{+26\%} higher performance than \textbf{APE} in larger data (100-150 trajectories), while remaining competitive with ML/DL methods (within 1.5\%). We can see that, varying \#trajectories with \#POI kept constant (bottom-left graph) has little effect on $F1$ score (compared to other graphs) as frequency of all the candidate paths are increased proportionally by repeating major patterns.

\shortname{}'s superior scalability stems from its specialized agent architecture. The \textit{Discovery Agent} efficiently filters historical data before ranking, reducing context size for downstream agents, compared to methods that process all trajectories simultaneously. When paths must be generated, the \textit{Synthesis Agent} constrains LLM outputs to edges present in historical data only, eliminating hallucinated connections that plague simple prompting approaches. This explicit validation maintains high $Traversability$ even as network sparsity increases. Meanwhile, the \textit{Popularity Ranking Agent}'s frequency-based scoring ensures that \shortname{} identifies truly popular paths and obtains high $F1$. In contrast, \textbf{LLM-A*} degrades rapidly because it relies on heuristic search without explicit edge constraints or popularity goal. \textbf{APE}'s iterative prompt engineering accumulates errors across feedback rounds in sparse networks, while \shortname{}'s modular design isolates and handles each sub-problem (discovery, ranking, synthesis, selection) independently.

%% file: figures/synthetic_data_eval.tex
\begin{figure*}[ht]
    \centering

    \resizebox{0.8\textwidth}{!}{%
        \begin{minipage}{\textwidth}
            \centering

            \includegraphics[width=0.5\textwidth]{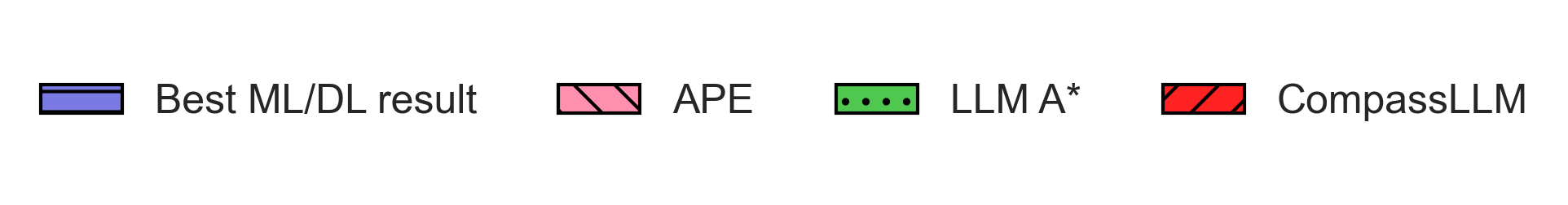}

            \begin{subfigure}[t]{0.49\textwidth}
                \centering
                \includegraphics[width=\textwidth]{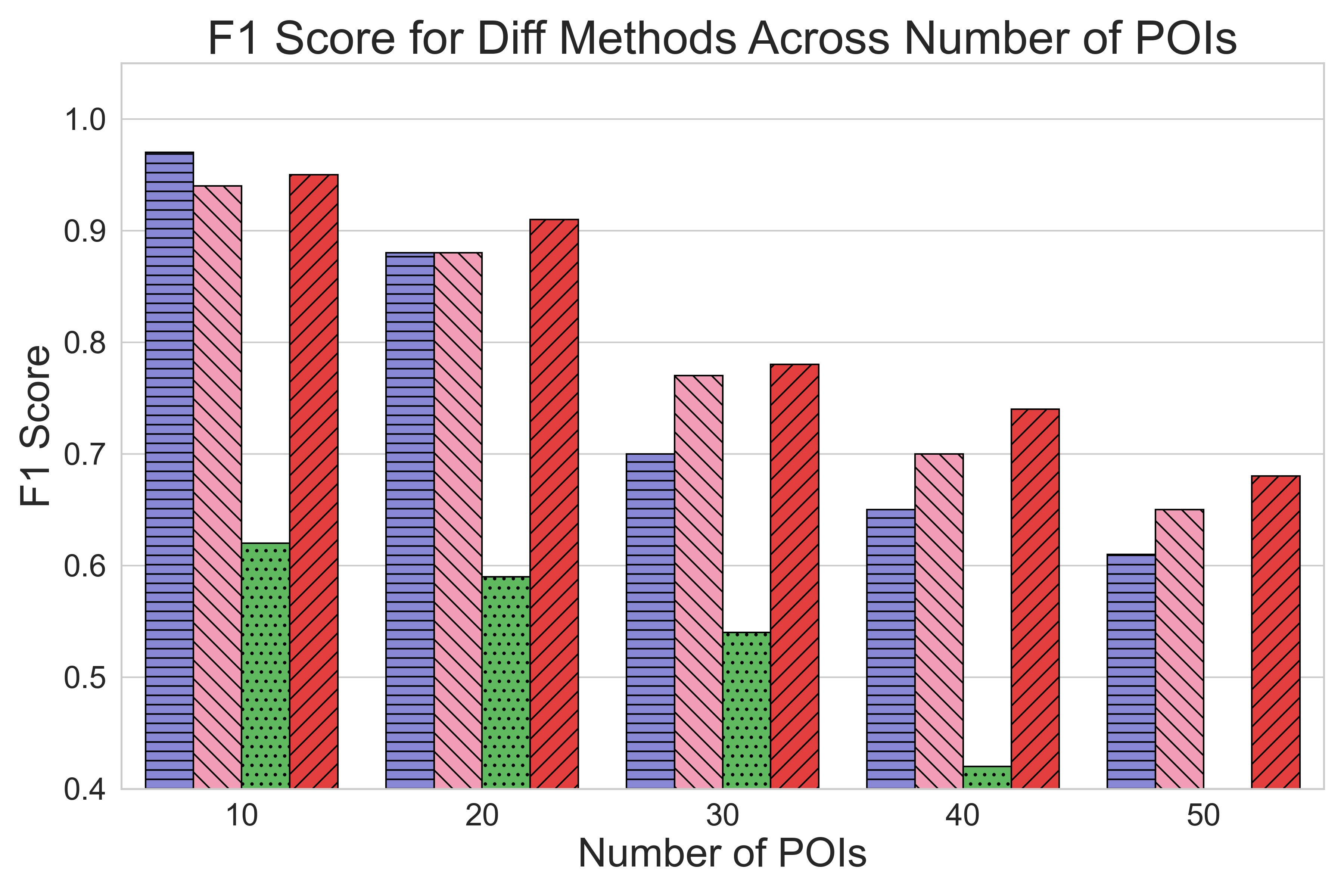}
                \label{fig:NodeVSF1}
            \end{subfigure}%
            \hfill%
            \begin{subfigure}[t]{0.49\textwidth}
                \centering
                \includegraphics[width=\textwidth]{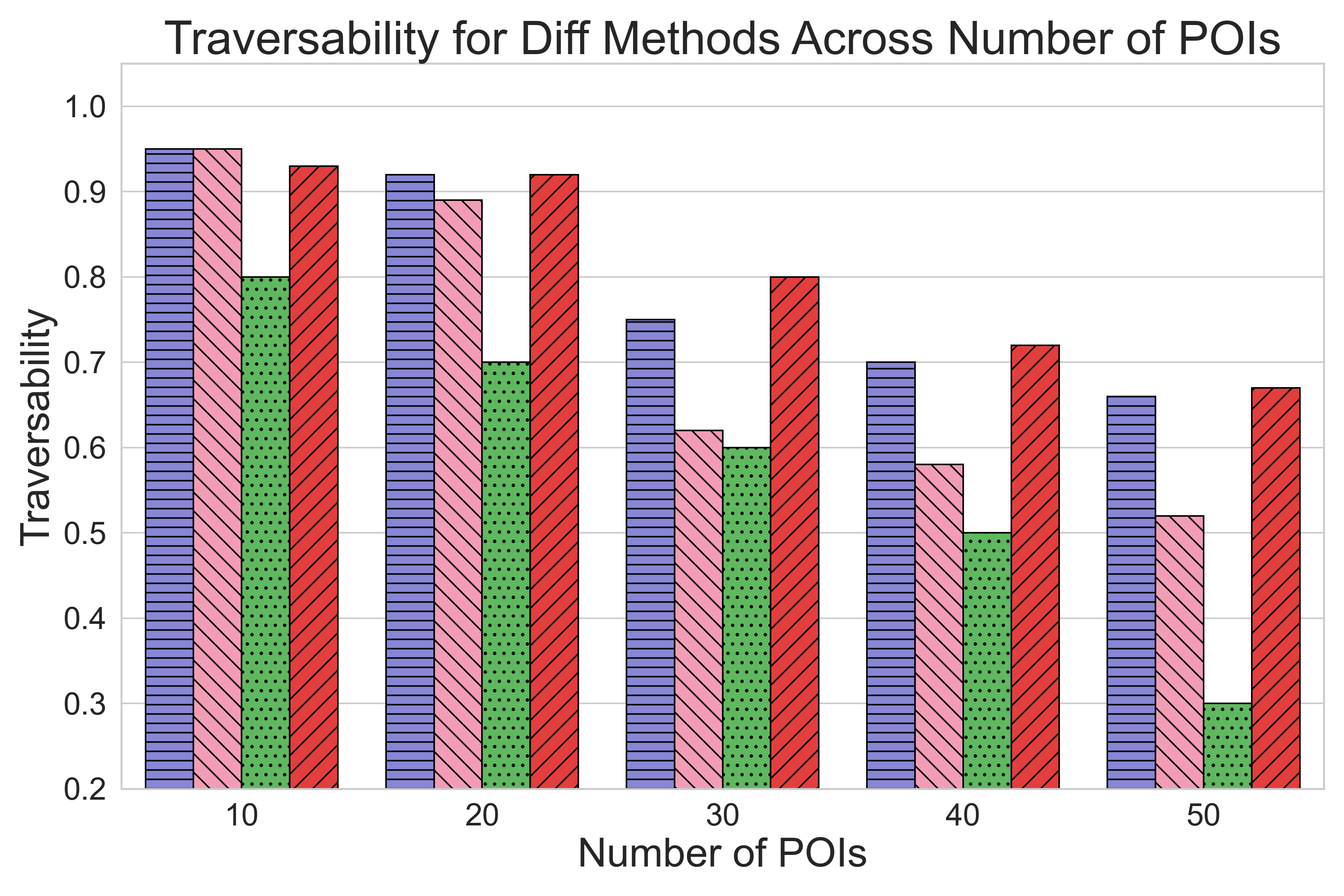}
                \label{fig:NodeVSReachability}
            \end{subfigure}

            \begin{subfigure}[t]{0.49\textwidth}
                \centering
                \includegraphics[width=\textwidth]{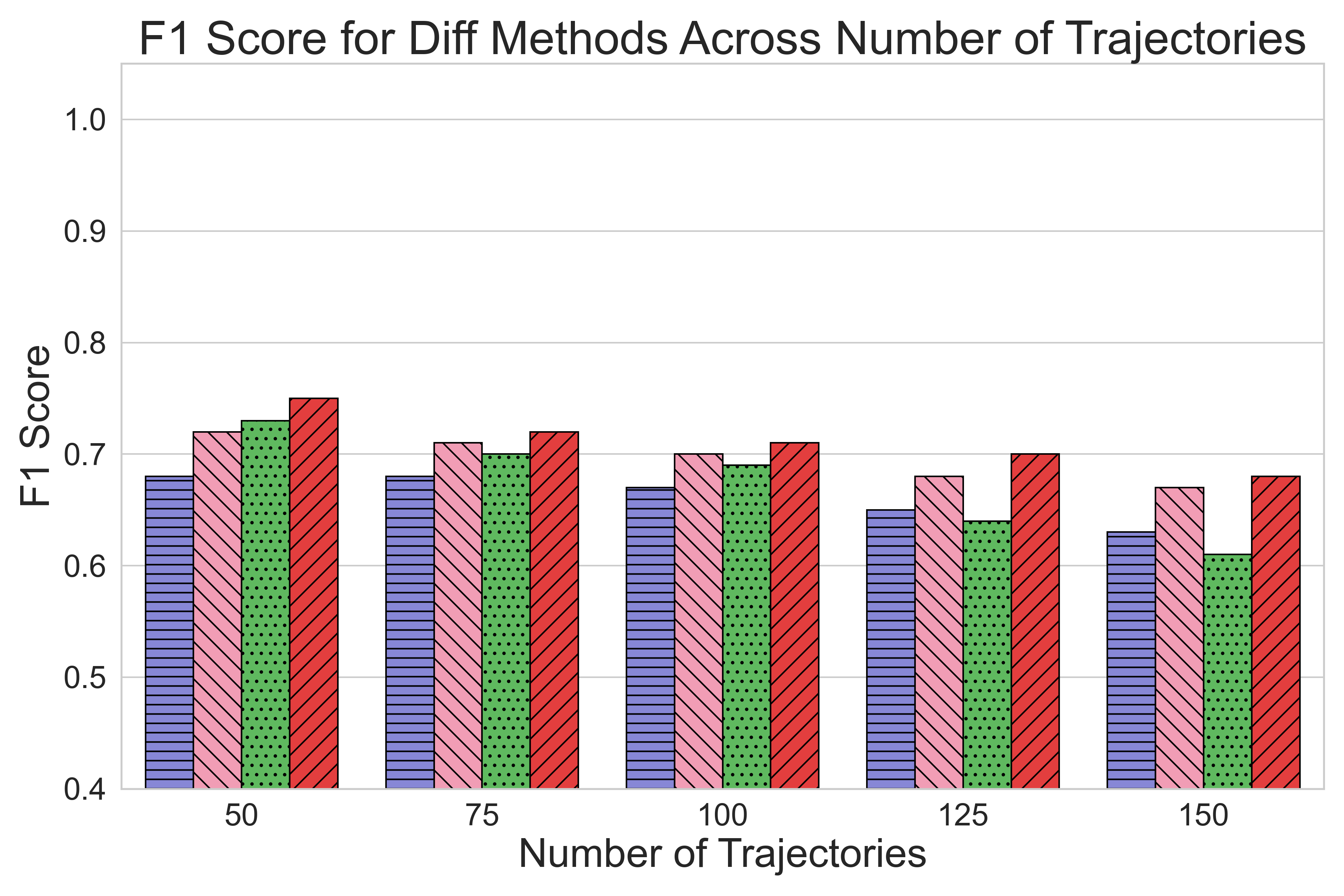}
                \label{fig:Image3}
            \end{subfigure}%
            \hfill%
            \begin{subfigure}[t]{0.49\textwidth}
                \centering
                \includegraphics[width=\textwidth]{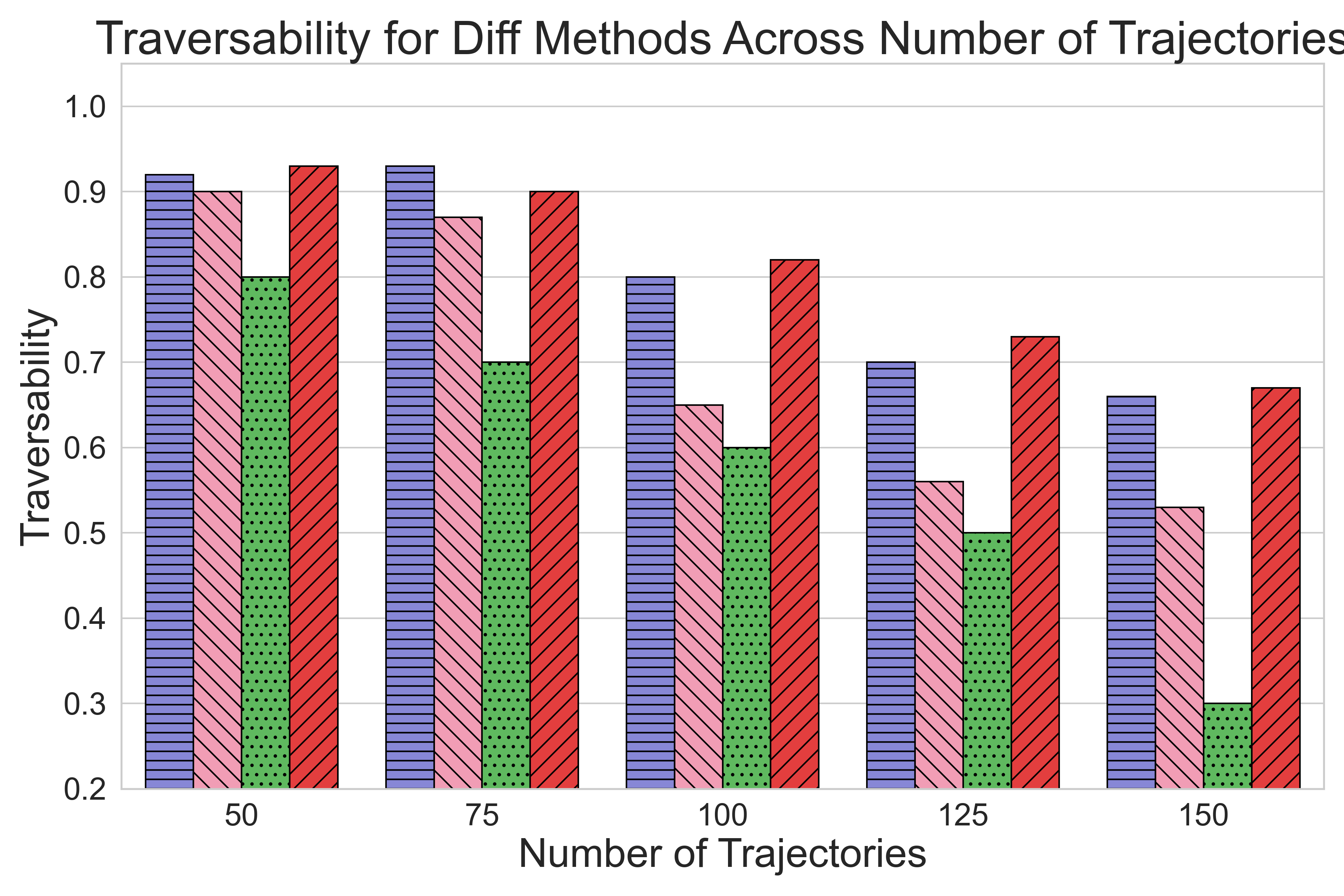}
                \label{fig:Image4}
            \end{subfigure}
        \end{minipage}%
    }

    \caption{
    Comparison among approaches on synthetic data.
    }
    \label{fig:combined_figures}
\end{figure*}

%% file: sections/conclusion.tex
In this work, we presented \shortname{}, a framework utilizing multiple agents for spatial reasoning to solve the Popular Path Problem. Our experiments show that \shortname{} is cost-effective, highly competitive in generating and often outperforming in finding paths, not to mention particularly effective under sparse data conditions. However, our results remain experimental, as \shortname{} may generate suboptimal or invalid paths in certain cases, and the inherent stochasticity of LLMs poses challenges in consistency. Our work bolsters LLMs' success in geo-spatial reasoning, inaugurates agentic approach, and could be extended in many directions. Future work can focus on incorporating more contextual data (e.g., time, user preference, and environmental factors), and enhancing models' reliability through improved prompt engineering and dynamic memory management to handle larger datasets effectively.

%% file: sections/limitations.tex
While \shortname{} demonstrates promising results in popular path discovery and synthesis, several limitations warrant consideration:

\textbf{Large-scale Data Handling:} \shortname{} faces challenges when processing extensive datasets. The current implementation struggles with input sizes exceeding 128,000 tokens, due to the context window limitations of existing LLMs, which limits its applicability to very large or complex spatial networks. Moreover, according to Table~\ref{tab:comparison_table} in Appendix, no further compression of \shortname{} prompts are feasible.

\textbf{Incompatibility with Internal Prompt Compression:} Many LLMs employ internal prompt compression techniques to manage large inputs efficiently. However, \shortname{}'s performance relies on the full, uncompressed prompt structure. Consequently, models that automatically compress prompts may not be compatible with our approach, potentially limiting the range of applicable LLMs.

\textbf{Prompt and Reasoning Chain Optimization:} While our current prompting strategy yields effective results, there may exist more optimized prompt structures or reasoning chains that could further enhance \shortname{}'s performance. The current implementation, while effective, may not represent the absolute optimal prompt design for all scenarios.

%% file: sections/z_appendix.tex
\section{\shortname{} Algorithm}
\label{app:Algorithm}

\begin{algorithm}[h]
\caption{\shortname{}}\label{alg:compassllm}
\footnotesize
\KwIn{$\mathcal{T}$ is a set of historical trajectories, $s$ is the start point, $d$ is the destination}
\KwOut{Popular path $\mathcal{R}$ from $s$ to $d$}

\BlankLine
$\mathcal{T}_{LLM} \leftarrow \text{PathDiscoveryAgent}(\mathcal{T}, s, d)$\;

\BlankLine
\tcp{No Candidates - GENERATE Block}
\uIf{$\mathcal{T}_{LLM} = \emptyset$}{
    $\texttt{EdgeRanks} \leftarrow \text{PopularityRankingAgent}(\mathcal{T},mode=edge)$\;

    \BlankLine
    \tcp{Update Candidates}
    $\mathcal{T}_{LLM} \leftarrow \text{PathSynthesisAgent}(s, d, \texttt{EdgeRanks})$
}

\BlankLine
$\texttt{POIRanks} \leftarrow \text{PopularityRankingAgent}(\mathcal{T},mode=poi)$\;

$\texttt{rankedPaths} \leftarrow \text{PathSelectionAgent}(\mathcal{T}_{LLM}, \texttt{POIRanks})$\;

\BlankLine
\tcp{Return best path}
$\mathcal{R} \leftarrow \texttt{rankedPaths}[0]$\;
\Return{$\mathcal{R}$}\;

\end{algorithm}

The full \shortname{} algorithm is outlined in Algorithm~\ref{alg:compassllm}.

\section{Synthetic Data Generation}
\label{app:SyntheticData}

\input{figures/steiner_tree}
\input{figures/comparison_real_synthetic}

\begin{table*}[!tb]
\centering

\resizebox{0.9\linewidth}{!}{%
\begin{tabular}{cccccccp{9cm}}
\toprule
\rotatebox{90}{\textbf{\shortstack{Path\\Discovery}}} & \rotatebox{90}{\textbf{\shortstack{Popularity\\Ranking}}} & \rotatebox{90}{\textbf{\shortstack{Path\\Synthesis}}} & \rotatebox{90}{\textbf{\shortstack{Path\\Selection}}} & \rotatebox{90}{$F1$} & \rotatebox{90}{$Traversability$} & \textbf{\shortstack{\textbf{Token}\\\textbf{Usage}}} & \textbf{Remarks} \\
\midrule
\checkmark & \checkmark & \checkmark & \checkmark & \textbf{0.80} & \textbf{0.95} & \textbf{11.8k} & Best performing \\
\hline
\texttimes & \checkmark & \checkmark & \checkmark & 0.63 & 0.91 & \shortstack{{}\\15.8k\\(increased)} & Always triggers GENERATE even when historical paths exist; slow and expensive; may generate worse paths than discovered ones; \textbf{shows necessity of conditional two-stage workflow} \\
\hline
\checkmark & \texttimes & \checkmark & \checkmark & 0.48 & 0.90 & 10.2k & Path Selection Agent has no popularity guidance for ranking candidates; essentially random/heuristic-based selection; \textbf{shows Popularity Ranking drives the system to Search for Popular paths effectively} \\
\hline
\checkmark & \checkmark & \texttimes & \checkmark & 0.80 & 0.42 & 9.5k & Only impacts GENERATE stage; greedy approach (pick most popular outgoing edge) creates locally optimal but globally invalid paths; can get stuck with no path to destination; \textbf{demonstrates need for global path-planning} \\
\hline
\checkmark & \checkmark & \checkmark & \texttimes & 0.77 & 0.95 & 10.6k & Simple arithmetic scoring captures core popularity logic; \textbf{shows LLM selection adds modest value; programmatic approach is competitive} \\
\bottomrule
\end{tabular}
}
\caption{Ablation Study Results}
\label{tab:ablation}
\end{table*}

\begin{table*}[!tb]
\centering
\resizebox{\linewidth}{!}{%
\begin{tabular}{lccccccccc}
\toprule
\textbf{Experiment} & \textbf{\shortstack{\shortname{}\\(GENERATE)}} & \textbf{\shortstack{Zero\\Shot}} & \textbf{CoT} & \textbf{\shortstack{Self-\\consistency}} & \textbf{ReAct} & \textbf{Reflexion} & \textbf{\shortstack{LLM\\A*}} & \textbf{APE} \\
\midrule
path with [[X]] POIs & 8 & 0 & 5 & 6 & 7 & 8 & 7 & 1 \\
path that goes through [[Y]] POI & 6 & 0 & 4 & 4 & 7 & 8 & 6 & 0 \\
path that avoids [[Z]] POI & 8 & 0 & 6 & 6 & 9 & 9 & 8 & 1 \\
\bottomrule
\end{tabular}
}
\caption{Comparison of constraint satisfaction across LLM-based approaches}
\label{tab:constraint}
\end{table*}

\input{tables/generated_actual_comparison}
\input{tables/reduction_reachability}

\subsection{Synthetic Data Generation Pipeline}

Our synthetic trajectory generation follows a four-step pipeline illustrated in Figure \ref{fig:steiner}, designed to create realistic spatial networks that mirror the structural properties of real-world transportation systems.

\textbf{Step 1: Graph Construction and Pivotal Node Identification.} We begin by constructing a base graph $G$ using the reverse-delete algorithm \cite{kruskal1956shortest} to generate a random connected network with controlled edge density. Within this network, we strategically identify pivotal nodes $V_s$ (depicted as yellow nodes) representing high-importance locations such as popular destinations, transportation hubs, or landmarks. Regular nodes (white nodes) serve as intermediate waypoints, creating a realistic spatial hierarchy.

\textbf{Step 2: Steiner Tree Construction.} Using approximation algorithms, we construct a Steiner tree $T_s$ that efficiently connects all pivotal nodes $V_s$. This tree structure, highlighted with pink/red edges, forms the backbone of our synthetic network and represents the primary "highway" system—the most efficient pathways between important locations that real-world travelers would naturally prefer.

\textbf{Step 3: Source-Destination Pair Selection.} We systematically select source node $s$ and destination node $d$ pairs from the network, ensuring diverse spatial configurations. This selection process considers both the connectivity patterns established by the Steiner tree and the spatial distribution of pivotal nodes to generate realistic travel scenarios.

\textbf{Step 4: Trajectory Synthesis.} We generate synthetic trajectories $\tau$ from source $s$ to destination $d$ using a hybrid approach: trajectories preferentially utilize Steiner tree edges (the highway network) for efficiency, while incorporating direct connections when they provide superior routes. This balances realistic movement behavior with network structure constraints.

\subsection{Validation Against Real-World Data}

Figure \ref{fig:RealFakeCompare} demonstrates that our synthetic data generation successfully captures the essential characteristics of real-world spatial patterns. The comparison reveals striking similarities between synthetic and real trajectory distributions:

\begin{itemize}
\item \textbf{Network Structure:} Both synthetic and real networks exhibit similar connectivity patterns, with certain paths (highlighted in red) experiencing significantly higher traffic than others (shown in grey).
\item \textbf{Frequency Distributions:} The density plots show comparable frequency distributions between synthetic and real data, with both exhibiting right-skewed distributions characteristic of real-world movement patterns where a few popular routes dominate usage.
\end{itemize}

Table \ref{tab:realdata_vs_syntheticdata} provides quantitative validation of our synthetic data generation, comparing the count of unique source-destination ground truth pairs between synthetic and real datasets. The "Result" column indicates generated unique pairs, while "Actual" represents corresponding real-world counts, demonstrating our method's ability to produce realistic trajectory volumes.

This systematic validation confirms that our synthetic trajectories exhibit authentic spatial characteristics while maintaining the structural properties necessary for effective LLM evaluation in spatial reasoning tasks. The density plots (where the area under each curve equals one) reveal that our synthetic data successfully replicates the underlying statistical patterns of real-world spatial movement, providing a reliable foundation for comprehensive algorithm evaluation.

This methodology enables us to generate synthetic trajectories that closely model actual spatial data, providing a nuanced representation of spatial interactions for comprehensive evaluation of LLM performance in spatial data analysis.

\input{figures/reachability_boxplot}

\begin{figure}[!htb]
    \centering
    \resizebox{\linewidth}{!}{%
    \includegraphics{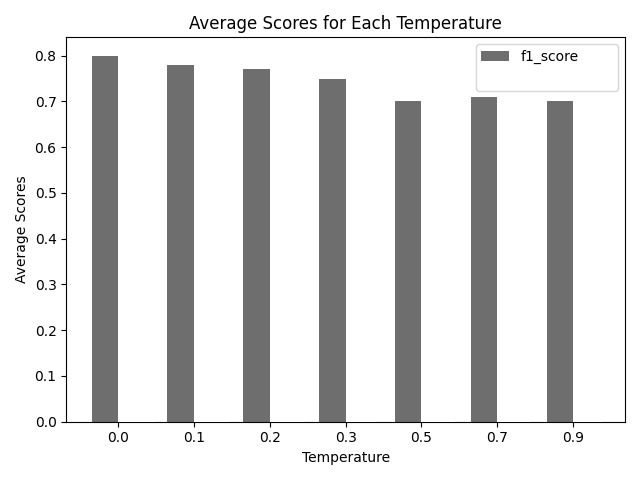}
    }
    \caption{Average F1 Score across all dataset against varying temperature}
    \label{fig:avg_temp_score}
\end{figure}

\section{Ablation}

An ablation study with dropping one agent at a time and replaced by algorithmic counterparts is presented at Table~\ref{tab:ablation}. Experiments are conducted with GPT o3 mini.

\section{Additional Results \& Trends}\label{app:additional_results}

\subsection{$F1$ with varying Temperature}\label{app:avg_temp}

Fig~\ref{fig:avg_temp_score} shows that LLMs generate better result with lower temperatures as a higher temperature can lead to hallucinating and a lower temperature gives out thoughtful answer. This led us to run all our experiments with temperature zero.

\subsection{Statistical Significance}\label{app:statistical_significance}
Across the 7 \textit{SEARCH} datasets, a one-sided Wilcoxon signed-rank test confirms that \shortname{} (GPT o3 mini) significantly outperforms every LLM and ML/DL baseline (p < 0.05 for all; p < 0.01 for all except LLM-A*). With GPT 4o, \shortname{} significantly outperforms most baselines but achieves parity with APE (mean difference +0.003, p = 0.38), consistent with our observation that gains are smaller in denser regimes. For \textit{GENERATE}, having only three datasets limits statistical power, so we refer to the variance analysis in Appendix \ref{app:reachability_consistency} and not significance testing.

\subsection{Variance of $Traversability$}\label{app:reachability_consistency}
A higher $traversability$ is always preferable but consistency is also necessary. Fig~\ref{fig:Reachability_Boxplot} shows that APE and \shortname{} are the only high performing and low variant ones.

\subsection{$Traversability$ with Reduction}
Selective Context \cite{li2023compressing} is a method that enhances the inference efficiency of LLMs by identifying and pruning redundancy in the input context to make the input more compact. Table \ref{tab:comparison_table} shows the comparison between \shortname{}, Selective Context 10\% and Selective Context 20\% in average reduction and $traversability$ across various datasets. This proves that \shortname{} prompts are already compact enough themselves and no level of reduction can improve/maintain the same performance. It validates the low-cost of \shortname{}.

\subsection{Constraint Handling}
We have experimented some graph prior constraints on the LLM based approaches - 10 experiment for each case. Table \ref{tab:constraint} shows a comparison of how many of the tasks each approach could solve properly.

\section{\shortname{} Example Prompts}\label{app:prompts}

As an adaptive framework, CompassLLM uses only the source-destination pair and historical paths as structured input \cite{gu2025structext}. While POI ID or Name could be used interchangeably in our solution \texttt{(25->16 or EdinCastle->Mound)}, doing so made little difference in our experiments. This indicates that no inherent knowledge in the LLMs about these locations was particularly useful in the process; only the orchestration itself drove the LLMs to achieve such results. Here are some example prompts for each agent.

\small
\ttfamily

\begin{promptBox}[]{}{Path Discovery Agent Prompt}{blue}
You are given a list of historical trajectories and a source-destination pair. Your task is to extract all candidate paths that connect the source to the destination.\\

\textbf{Historical Trajectories:}\\
Calton Hill -> National Monument -> Arthur's Seat; \\
Edinburgh Castle -> Royal Mile -> National Museum of Scotland -> University of Edinburgh -> Calton Hill -> Royal Botanic Garden -> Edinburgh Zoo -> Greyfriars Kirkyard -> Camera Obscura -> Scottish Parliament -> Nelson Monument -> Dynamic Earth; \\
Princes Street Gardens -> Palace of Holyroodhouse -> National Museum of Scotland -> Scott Monument -> St Giles' Cathedral -> Dynamic Earth; \\
Edinburgh Castle -> Palace of Holyroodhouse -> St Giles' Cathedral -> Greyfriars Kirkyard -> Dynamic Earth; \\
\dots \\
Royal Mile -> Arthur's Seat -> Nelson Monument; \\
Arthur's Seat -> National Museum of Scotland -> Edinburgh Zoo; \\
Edinburgh Castle -> National Museum of Scotland -> Edinburgh Zoo;\\

\textbf{Source:} Royal Botanic Garden \\
\textbf{Destination:} Arthur's Seat\\

---------------\\

\textbf{Important:} Your response must follow the following JSON format:\\

\{\\
\mytab "identification\_process": "Explain how we can identify existing paths from historical data that contain both source and destination.",\\
\mytab "candidate\_paths": "Retrieve all possible routes connecting the source and destination from the historical data."\\
\}\\

\end{promptBox}

\begin{promptBox}[]{}{Popularity Ranking Agent Prompt}{blue}
You are given historical trajectory data. Your task is to analyze and rank all \{POIs $|$ POI Pairs (Edges)\} based on their popularity.\\

\textbf{Historical Trajectories:}\\
Calton Hill -> National Monument -> Arthur's Seat; \\
Edinburgh Castle -> Royal Mile -> National Museum of Scotland -> University of Edinburgh -> Calton Hill -> Royal Botanic Garden -> Edinburgh Zoo -> Greyfriars Kirkyard -> Camera Obscura -> Scottish Parliament -> Nelson Monument -> Dynamic Earth; \\
Princes Street Gardens -> Palace of Holyroodhouse -> National Museum of Scotland -> Scott Monument -> St Giles' Cathedral -> Dynamic Earth; \\
Edinburgh Castle -> Palace of Holyroodhouse -> St Giles' Cathedral -> Greyfriars Kirkyard -> Dynamic Earth; \\
\dots \\
Royal Mile -> Arthur's Seat -> Nelson Monument; \\
Arthur's Seat -> National Museum of Scotland -> Edinburgh Zoo; \\
Edinburgh Castle -> National Museum of Scotland -> Edinburgh Zoo;\\

---------------\\

\textbf{Important:} Your response must follow the following JSON format:\\

\textbf{For POI Mode:}\\

\{\\
\mytab "calculation\_method": "Step by step explain how the popularity of various points of interest (POIs) can be analyzed using historical trajectory data.",\\
\mytab "ranking\_analysis": "Analyze the ranking the POIs based on their frequency of visits or interactions in the dataset and provide the ranking.",\\
\mytab "poi\_rank": "Give out the POIs in the ascending order of rank."\\
\}\\
\\

\textbf{For Edge Mode:}\\

\{\\
\mytab "extracted\_edges": "Extract edges that can be found from the trajectory data.",\\
\mytab "analysis\_method": "How to analyze edge frequency and find the most popular edges.",\\
\mytab "edge\_rank": "Give out the edges in (POI1, POI2) format in descending order of their popularity."\\
\}\\

\end{promptBox}

\begin{promptBox}[]{}{Path Synthesis Agent Prompt}{blue}
You are given edge popularity rankings. For a source-destination pair with no existing path in historical data, generate candidate paths.\\

\textbf{Source:} Royal Botanic Garden \\
\textbf{Destination:} Arthur's Seat\\

\textbf{Edge Popularity Ranking:} (Dynamic Earth, Scott Monument), (Calton Hill, Royal Botanic Garden), (Royal Mile, National Museum of Scotland), (Scott Monument, Calton Hill), (Edinburgh Castle, Royal Mile), (National Museum of Scotland, University of Edinburgh), (Calton Hill, National Monument), (National Monument, Arthur's Seat)...\\

---------------\\

\textbf{Important:} Your response must follow the following JSON format:\\

\{\\
\mytab "generation\_strategy": "Explain the strategy for combining popular edges to create realistic paths.",\\
\mytab "generated\_paths": "Generate possible path candidates using edge popularity rankings."\\
\}\\

\end{promptBox}

\begin{promptBox}[]{}{Path Selection Agent Prompt}{blue}
You are given candidate paths extracted from historical data and POI popularity rankings. Your task is to rank these paths based on their popularity.\\

\textbf{Candidate Paths:}\\
Royal Botanic Garden -> Calton Hill -> National Monument -> Arthur's Seat; \\
Royal Botanic Garden -> Edinburgh Zoo -> National Museum of Scotland -> University of Edinburgh -> Royal Mile -> Arthur's Seat; \\
Royal Botanic Garden -> Princes Street Gardens -> Palace of Holyroodhouse -> Arthur's Seat;\\

\textbf{POI Popularity Ranking:} National Museum of Scotland, Scott Monument, Edinburgh Zoo, St Giles' Cathedral, Royal Mile, Greyfriars Kirkyard, Royal Botanic Garden, Calton Hill, Arthur's Seat, Palace of Holyroodhouse, Edinburgh Castle, Camera Obscura, University of Edinburgh, National Monument\\

---------------\\

\textbf{Important:} Your response must follow the following JSON format:\\

\{\\
\mytab "evaluation\_method": "Think step by step on how to evaluate paths using POI popularity rankings and retrieve the best path.",\\
\mytab "ranked\_paths": "Rank all candidate paths based on the popularity of POIs they traverse."\\
\}\\

\end{promptBox}

%% file: figures/steiner_tree.tex
\begin{figure}[h]
    \centering
    \includegraphics[width=0.6\linewidth]{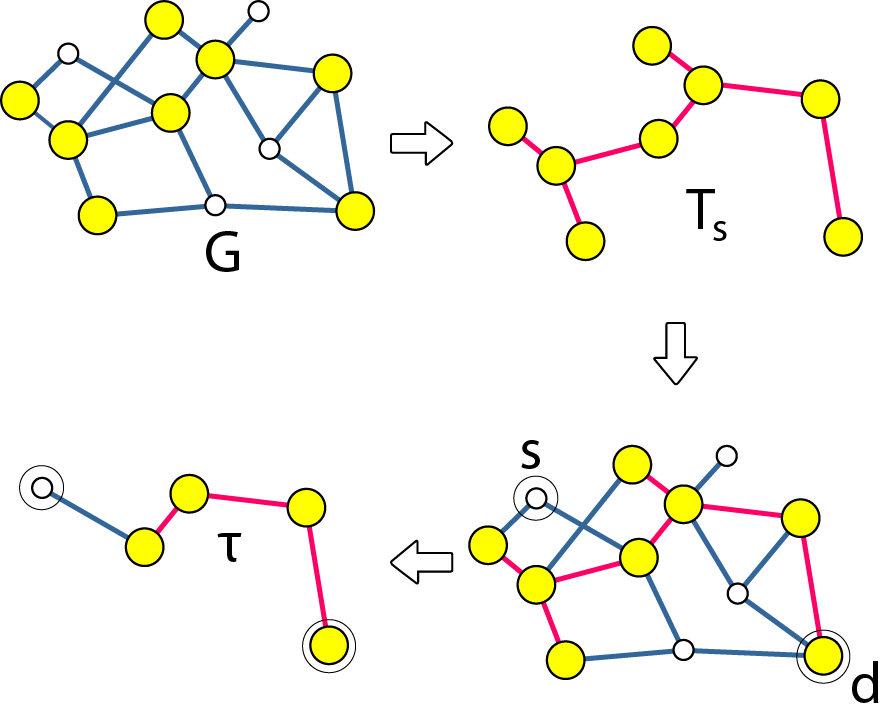}
    \caption{Synthetic Data Generation Process}
    \label{fig:steiner}
\end{figure}

%% file: figures/comparison_real_synthetic.tex
\begin{figure}[htb]
    \centering
    \resizebox{\linewidth}{!}{%
    \includegraphics{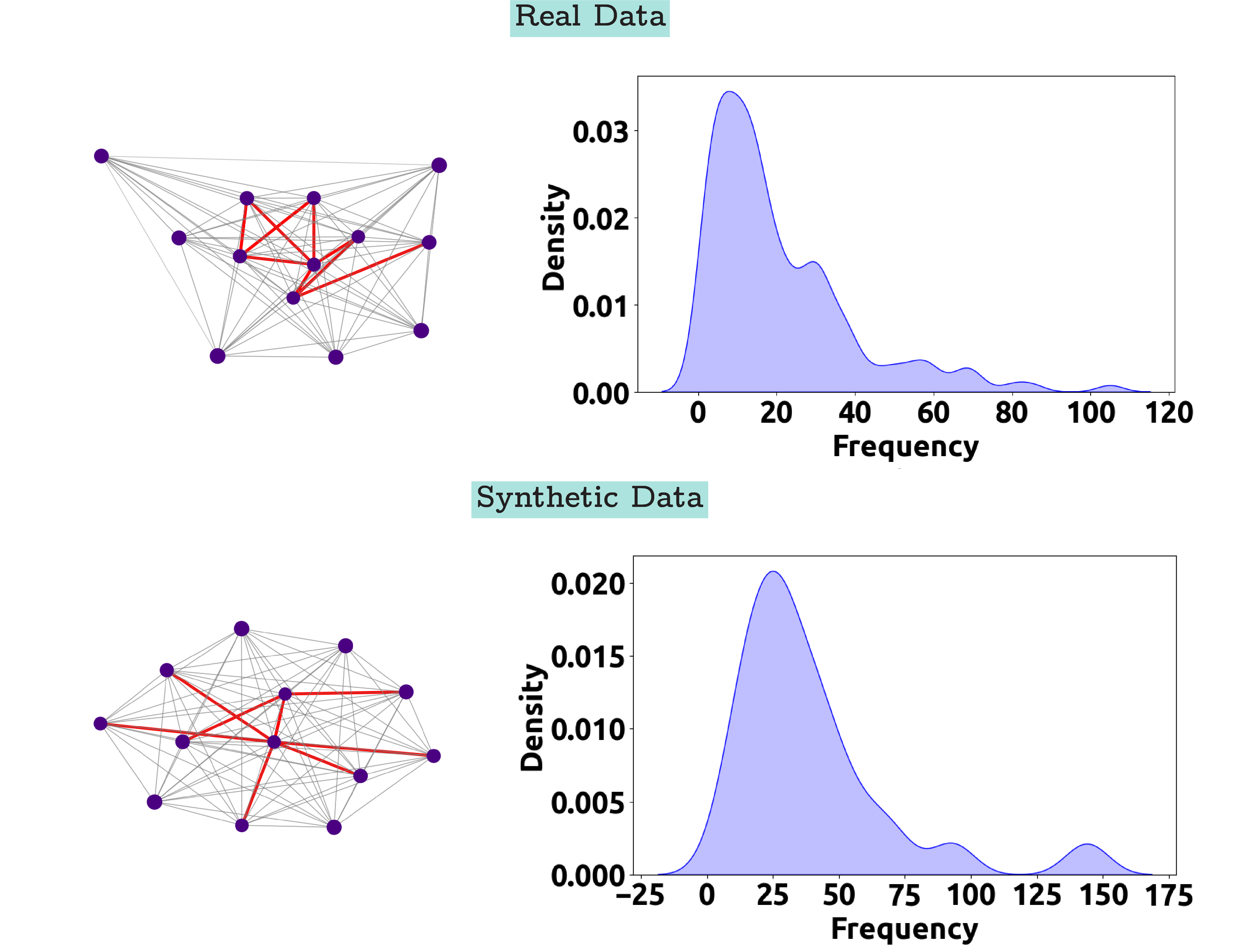}
    }
    \caption{Comparison of Real and Synthetic Data}
    \label{fig:RealFakeCompare}
\end{figure}

%% file: tables/generated_actual_comparison.tex
\begin{table}[h]
\centering
\small
\begin{tabularx}{\linewidth}{l*{5}{>{\centering\arraybackslash}X}}
\toprule
\textbf{Location} & \textbf{Nodes} & \textbf{Density} & \textbf{Routes} & \textbf{Result} & \textbf{Actual} \\
\midrule
Epcot & 17 & 0.2 & 1248 & 212 & 207 \\
Disneyland & 31 & 0.4 & 2792 & 681 & 618 \\
DisHolly & 13 & 0.4 & 901 & 107 & 134 \\
\bottomrule
\end{tabularx}
\caption{Comparison of synthetic and real-world data across multiple locations. The table presents key characteristics of the generated spatial datasets, including the number of nodes, spatial density, and the number of unique routes. The "Result" column denotes the number of unique source-destination pairs generated by our synthetic trajectory model, while the "Actual" column provides the corresponding real-world counts.}
\label{tab:realdata_vs_syntheticdata}
\end{table}

%% file: tables/reduction_reachability.tex
\begin{table*}[tb]
    \centering
    \small
        \begin{tabular}{llcccccc}
        \toprule
        \textbf{Dataset} & \textbf{Metric} & \textbf{\shortname{}} & \textbf{Selective Context} & \textbf{Selective Context} \\
        & & \textbf{GENERATE} & \textbf{(10\%)} & \textbf{(20\%)} \\
        \midrule
        \multirow{2}{*}{\textbf{Epcot}}
         & Reduction & 0 & 16\% & 24\% \\
         & $Traversability$ & 0.84 & 0.15 & 0 \\
        \midrule
        \multirow{2}{*}{\textbf{CaliAdv}}
         & Reduction & 0 & 14\% & 22\% \\
         & $Traversability$ & 0.91 & 0.12 & 0 \\
        \midrule
        \multirow{2}{*}{\textbf{Disland}}
         & Reduction & 0 & 13\% & 20\% \\
         & $Traversability$ & 0.96 & 0.10 & 0.1 \\
        \midrule
        \multirow{2}{*}{\textbf{Average}}
         & Reduction & 0 & 13.75\% & 21\% \\
         & $Traversability$ & 0.80 & 0.14 & 0.05 \\
        \bottomrule
        \end{tabular}
    \caption{Comparison of Reduction and $Traversability$ across various datasets and techniques.}
    \label{tab:comparison_table}
\end{table*}

%% file: figures/reachability_boxplot.tex
\begin{figure}[]
    \centering
    \resizebox{\linewidth}{!}{%
    \includegraphics{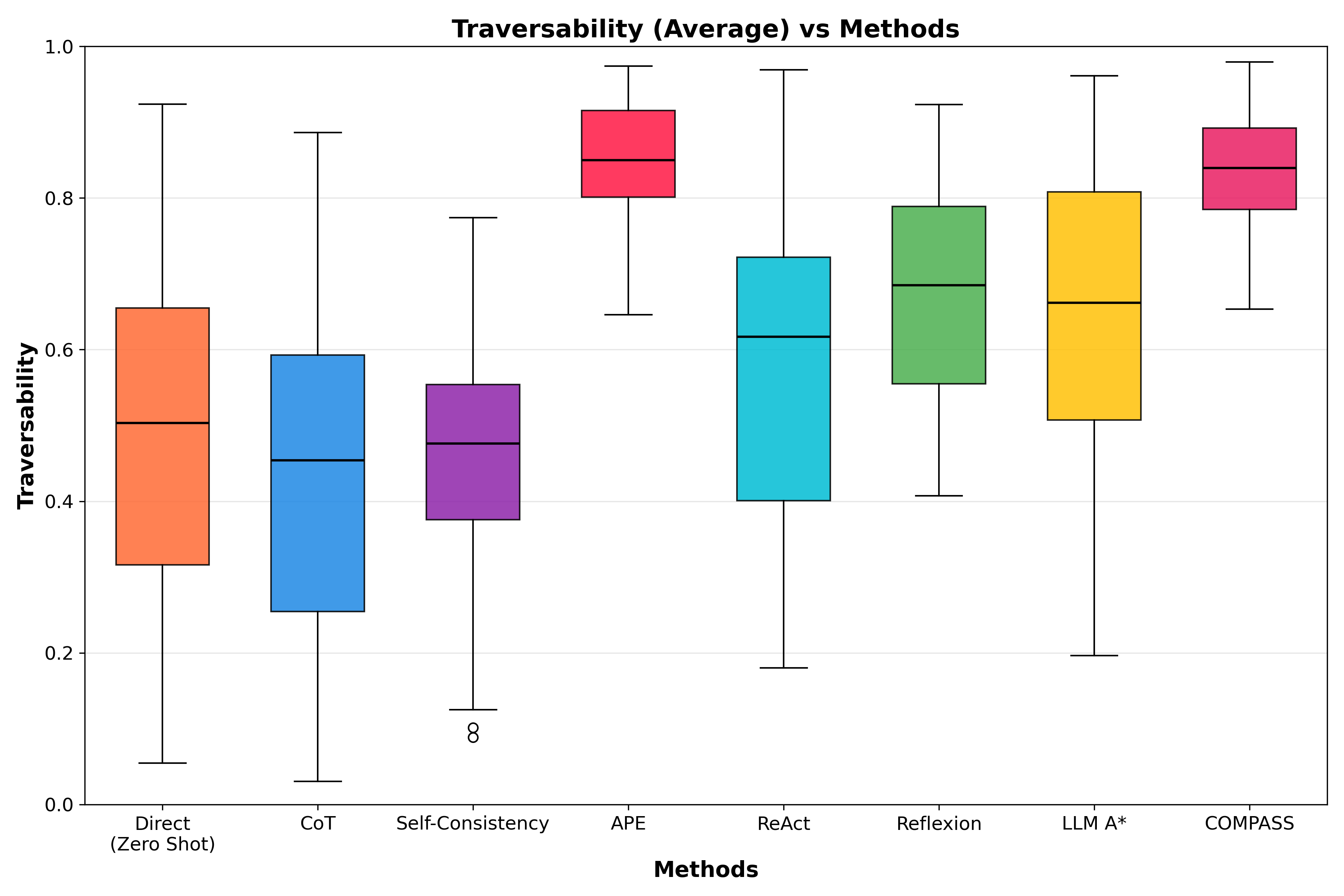}
    }
    \caption{Consistency Across Prompting Techniques: This box plot illustrates the variation in $Traversability$ scores across different prompting techniques. Lower variance is preferable, as it indicates more consistent results.}
    \label{fig:Reachability_Boxplot}
\end{figure}

%% file: custom.bib
@article{meng2024llm,
  title={LLM-A*: Large Language Model Enhanced Incremental Heuristic Search on Path Planning},
  author={Meng, Silin and Wang, Yiwei and Yang, Cheng-Fu and Peng, Nanyun and Chang, Kai-Wei},
  journal={arXiv preprint arXiv:2407.02511},
  year={2024}
}

@article{aghzal2023can,
  title={Can large language models be good path planners? a benchmark and investigation on spatial-temporal reasoning},
  author={Aghzal, Mohamed and Plaku, Erion and Yao, Ziyu},
  journal={arXiv preprint arXiv:2310.03249},
  year={2023}
}

@article{laskar2024systematic,
  title={A systematic survey and critical review on evaluating large language models: Challenges, limitations, and recommendations},
  author={Laskar, Md Tahmid Rahman and Alqahtani, Sawsan and Bari, M Saiful and Rahman, Mizanur and Khan, Mohammad Abdullah Matin and Khan, Haidar and Jahan, Israt and Bhuiyan, Amran and Tan, Chee Wei and Parvez, Md Rizwan and others},
  journal={arXiv preprint arXiv:2407.04069},
  year={2024}
}

@article{jain2021neuromlr,
  title={Neuromlr: Robust \& reliable route recommendation on road networks},
  author={Jain, Jayant and Bagadia, Vrittika and Manchanda, Sahil and Ranu, Sayan},
  journal={Advances in Neural Information Processing Systems},
  volume={34},
  pages={22070--22082},
  year={2021}
}

@article{kruskal1956shortest,
  title={On the shortest spanning subtree of a graph and the traveling salesman problem},
  author={Kruskal, Joseph B},
  journal={Proceedings of the American Mathematical society},
  volume={7},
  number={1},
  pages={48--50},
  year={1956}
}

@inproceedings{chen2016learning,
  title={Learning points and routes to recommend trajectories},
  author={Chen, Dawei and Ong, Cheng Soon and Xie, Lexing},
  booktitle={The Conference on Information and Knowledge Management (CIKM)},
  pages={2227--2232},
  year={2016}
}

@inproceedings{wei2012constructing,
  title={Constructing popular routes from uncertain trajectories},
  author={Wei, Ling-Yin and Zheng, Yu and Peng, Wen-Chih},
  booktitle={SIGKDD},
  pages={195--203},
  year={2012}
}

@inproceedings{banerjee2014inferring,
  title={Inferring uncertain trajectories from partial observations},
  author={Banerjee, Prithu and Ranu, Sayan and Raghavan, Sriram},
  booktitle={ICDM},
  pages={30--39},
  year={2014}
}

@inproceedings{chen2011discovering,
  title={Discovering popular routes from trajectories},
  author={Chen, Z. and Shen, H. T. and Zhou, X.},
  booktitle={ICDE},
  pages={900--911},
  year={2011}
}

@inproceedings{li2020spatial,
  title={Spatial transition learning on road networks with deep probabilistic models},
  author={Li, Xiucheng and Cong, Gao and Cheng, Yun},
  booktitle={ICDE},
  pages={349--360},
  year={2020}
}

@inproceedings{wang2019empowering,
  title={Empowering A* search algorithms with neural networks for personalized route recommendation},
  author={Wang, Jingyuan and Wu, Ning and Zhao, Wayne Xin and Peng, Fanzhang and Lin, Xin},
  booktitle={KDD},
  pages={539--547},
  year={2019}
}

@inproceedings{wu2017modeling,
  title={Modeling trajectories with recurrent neural networks},
  author={Wu, Hao and Chen, Ziyang and Sun, Weiwei and Zheng, Baihua and Wang, Wei},
  booktitle={IJCAI},
  pages={3083--3090},
  year={2017}
}

@inproceedings{wu2016probabilistic,
  title={Probabilistic robust route recovery with spatio-temporal dynamics},
  author={Wu, Hao and Mao, Jiangyun and Sun, Weiwei and Zheng, Baihua and Zhang, Hanyuan and Chen, Ziyang and Wang, Wei},
  booktitle={KDD},
  pages={1915--1924},
  year={2016}
}

@article{yang2018fast,
  title={Fast map matching, an algorithm integrating hidden markov model with precomputation},
  author={Yang, Can and Gidofalvi, Gyozo},
  journal={International Journal of Geographical Information Science},
  volume={32},
  number={3},
  pages={547--570},
  year={2018}
}

@inproceedings{zheng2012reducing,
  title={Reducing uncertainty of low-sampling-rate trajectories},
  author={Zheng, K. and Zheng, Y. and Xie, X. and Zhou, X.},
  booktitle={ICDE},
  pages={1144--1155},
  year={2012}
}

@misc{wei2023chainofthought,
      title={Chain-of-Thought Prompting Elicits Reasoning in Large Language Models}, 
      author={Jason Wei and Xuezhi Wang and Dale Schuurmans and Maarten Bosma and Brian Ichter and Fei Xia and Ed Chi and Quoc Le and Denny Zhou},
      year={2023},
      eprint={2201.11903},
      archivePrefix={arXiv},
      primaryClass={id='cs.CL' full_name='Computation and Language' is_active=True alt_name='cmp-lg' in_archive='cs' is_general=False description='Covers natural language processing. Roughly includes material in ACM Subject Class I.2.7. Note that work on artificial languages (programming languages, logics, formal systems) that does not explicitly address natural-language issues broadly construed (natural-language processing, computational linguistics, speech, text retrieval, etc.) is not appropriate for this area.'}
}

@misc{wang2023selfconsistency,
      title={Self-Consistency Improves Chain of Thought Reasoning in Language Models}, 
      author={Xuezhi Wang and Jason Wei and Dale Schuurmans and Quoc Le and Ed Chi and Sharan Narang and Aakanksha Chowdhery and Denny Zhou},
      year={2023},
      eprint={2203.11171},
      archivePrefix={arXiv},
      primaryClass={id='cs.CL' full_name='Computation and Language' is_active=True alt_name='cmp-lg' in_archive='cs' is_general=False description='Covers natural language processing. Roughly includes material in ACM Subject Class I.2.7. Note that work on artificial languages (programming languages, logics, formal systems) that does not explicitly address natural-language issues broadly construed (natural-language processing, computational linguistics, speech, text retrieval, etc.) is not appropriate for this area.'}
}

@inproceedings{song2023llm,
  title={Llm-planner: Few-shot grounded planning for embodied agents with large language models},
  author={Song, Chan Hee and Wu, Jiaman and Washington, Clayton and Sadler, Brian M and Chao, Wei-Lun and Su, Yu},
  booktitle={Proceedings of the IEEE/CVF International Conference on Computer Vision},
  pages={2998--3009},
  year={2023}
}

@article{gundawar2024robust,
  title={Robust Planning with LLM-Modulo Framework: Case Study in Travel Planning},
  author={Gundawar, Atharva and Verma, Mudit and Guan, Lin and Valmeekam, Karthik and Bhambri, Siddhant and Kambhampati, Subbarao},
  journal={arXiv preprint arXiv:2405.20625},
  year={2024}
}

@article{sharan2023llm,
  title={Llm-assist: Enhancing closed-loop planning with language-based reasoning},
  author={Sharan, SP and Pittaluga, Francesco and Chandraker, Manmohan and others},
  journal={arXiv preprint arXiv:2401.00125},
  year={2023}
}

@article{kambhampati2024llms,
  title={LLMs Can't Plan, But Can Help Planning in LLM-Modulo Frameworks},
  author={Kambhampati, Subbarao and Valmeekam, Karthik and Guan, Lin and Stechly, Kaya and Verma, Mudit and Bhambri, Siddhant and Saldyt, Lucas and Murthy, Anil},
  journal={arXiv preprint arXiv:2402.01817},
  year={2024}
}

@article{chen2024exploring,
  title={Exploring the potential of large language models (llms) in learning on graphs},
  author={Chen, Zhikai and Mao, Haitao and Li, Hang and Jin, Wei and Wen, Hongzhi and Wei, Xiaochi and Wang, Shuaiqiang and Yin, Dawei and Fan, Wenqi and Liu, Hui and others},
  journal={ACM SIGKDD Explorations Newsletter},
  volume={25},
  number={2},
  pages={42--61},
  year={2024},
  publisher={ACM New York, NY, USA}
}

@article{ye2023natural,
  title={Natural language is all a graph needs},
  author={Ye, Ruosong and Zhang, Caiqi and Wang, Runhui and Xu, Shuyuan and Zhang, Yongfeng},
  journal={arXiv preprint arXiv:2308.07134},
  year={2023}
}

@article{wang2024can,
  title={Can language models solve graph problems in natural language?},
  author={Wang, Heng and Feng, Shangbin and He, Tianxing and Tan, Zhaoxuan and Han, Xiaochuang and Tsvetkov, Yulia},
  journal={Advances in Neural Information Processing Systems},
  volume={36},
  year={2024}
}

@article{tian2023effective,
  title={Effective and Efficient Route Planning Using Historical Trajectories on Road Networks},
  author={Tian, Wei and Shi, Jieming and Luo, Siqiang and Li, Hui and Xie, Xike and Zou, Yuanhang},
  journal={Proceedings of the VLDB Endowment},
  volume={16},
  number={10},
  pages={2512--2524},
  year={2023},
  publisher={VLDB Endowment}
}

@article{rashid2023deepalttrip,
  title={DeepAltTrip: Top-K alternative itineraries for trip recommendation},
  author={Rashid, Syed Md Mukit and Ali, Mohammed Eunus and Cheema, Muhammad Aamir},
  journal={IEEE Transactions on Knowledge and Data Engineering},
  volume={35},
  number={9},
  pages={9433--9447},
  year={2023},
  publisher={IEEE}
}

@inproceedings{
yao2023react,
title={ReAct: Synergizing Reasoning and Acting in Language Models},
author={Shunyu Yao and Jeffrey Zhao and Dian Yu and Nan Du and Izhak Shafran and Karthik R Narasimhan and Yuan Cao},
booktitle={The Eleventh International Conference on Learning Representations },
year={2023}
}

@inproceedings{robins2000improved,
  title={Improved steiner tree approximation in graphs.},
  author={Robins, Gabriel and Zelikovsky, Alexander},
  booktitle={SODA},
  pages={770--779},
  year={2000}
}

@misc{shinn2023reflexionlanguageagentsverbal,
      title={Reflexion: Language Agents with Verbal Reinforcement Learning}, 
      author={Noah Shinn and Federico Cassano and Edward Berman and Ashwin Gopinath and Karthik Narasimhan and Shunyu Yao},
      year={2023},
      eprint={2303.11366},
      archivePrefix={arXiv},
      primaryClass={cs.AI},
      url={https://arxiv.org/abs/2303.11366}, 
}

@article{tam2024let,
  title={Let me speak freely? a study on the impact of format restrictions on performance of large language models},
  author={Tam, Zhi Rui and Wu, Cheng-Kuang and Tsai, Yi-Lin and Lin, Chieh-Yen and Lee, Hung-yi and Chen, Yun-Nung},
  journal={arXiv preprint arXiv:2408.02442},
  year={2024}
}

@article{10.1007/s10115-017-1056-y,
author = {Lim, Kwan Hui and Chan, Jeffrey and Leckie, Christopher and Karunasekera, Shanika},
title = {Personalized trip recommendation for tourists based on user interests, points of interest visit durations and visit recency},
year = {2018},
issue_date = {February  2018},
publisher = {Springer-Verlag},
address = {Berlin, Heidelberg},
volume = {54},
number = {2},
issn = {0219-1377},
url = {https://doi.org/10.1007/s10115-017-1056-y},
doi = {10.1007/s10115-017-1056-y},
journal = {Knowl. Inf. Syst.},
month = feb,
pages = {375–406},
numpages = {32}
}

@inproceedings{36046891aa77462298f21dacee3279a6,
title = "Learning points and routes to recommend trajectories",
author = "Dawei Chen and Ong, {Cheng Soon} and Lexing Xie",
note = "Publisher Copyright: {\textcopyright} 2016 ACM.; 25th ACM International Conference on Information and Knowledge Management, CIKM 2016 ; Conference date: 24-10-2016 Through 28-10-2016",
year = "2016",
month = oct,
day = "24",
doi = "10.1145/2983323.2983672",
language = "English",
series = "International Conference on Information and Knowledge Management, Proceedings",
publisher = "Association for Computing Machinery",
pages = "2227--2232",
booktitle = "CIKM 2016 - Proceedings of the 2016 ACM Conference on Information and Knowledge Management",
}

@article{zhou2022large,
  title={Large language models are human-level prompt engineers},
  author={Zhou, Yongchao and Muresanu, Andrei Ioan and Han, Ziwen and Paster, Keiran and Pitis, Silviu and Chan, Harris and Ba, Jimmy},
  journal={arXiv preprint arXiv:2211.01910},
  year={2022}
}

@inproceedings{zheng2011urban,
  title={Urban computing with taxicabs},
  author={Zheng, Yu and Liu, Yanchi and Yuan, Jing and Xie, Xing},
  booktitle={Proceedings of the 13th international conference on Ubiquitous computing},
  pages={89--98},
  year={2011}
}

@article{li2010swarm,
  title={Swarm: Mining relaxed temporal moving object clusters},
  author={Li, Zhenhui and Ding, Bolin and Han, Jiawei and Kays, Roland},
  journal={Proceedings of the VLDB Endowment},
  volume={3},
  number={1-2},
  pages={723--734},
  year={2010},
  publisher={VLDB Endowment}
}

@inproceedings{yuan2011driving,
  title={Driving with knowledge from the physical world},
  author={Yuan, Jing and Zheng, Yu and Xie, Xing and Sun, Guangzhong},
  booktitle={Proceedings of the 17th ACM SIGKDD international conference on Knowledge discovery and data mining},
  pages={316--324},
  year={2011}
}

@inproceedings{yuan2010t,
  title={T-drive: driving directions based on taxi trajectories},
  author={Yuan, Jing and Zheng, Yu and Zhang, Chengyang and Xie, Wenlei and Xie, Xing and Sun, Guangzhong and Huang, Yan},
  booktitle={Proceedings of the 18th SIGSPATIAL International conference on advances in geographic information systems},
  pages={99--108},
  year={2010}
}

@inproceedings{luo2013finding,
  title={Finding time period-based most frequent path in big trajectory data},
  author={Luo, Wuman and Tan, Haoyu and Chen, Lei and Ni, Lionel M},
  booktitle={Proceedings of the 2013 ACM SIGMOD international conference on management of data},
  pages={713--724},
  year={2013}
}

@misc{meng2024llmalargelanguagemodel,
      title={LLM-A*: Large Language Model Enhanced Incremental Heuristic Search on Path Planning}, 
      author={Silin Meng and Yiwei Wang and Cheng-Fu Yang and Nanyun Peng and Kai-Wei Chang},
      year={2024},
      eprint={2407.02511},
      archivePrefix={arXiv},
      primaryClass={cs.RO},
      url={https://arxiv.org/abs/2407.02511}, 
}

@INPROCEEDINGS{li2024research,
  author={Li, Bohang and Zhang, Kai and Sun, Yiping and Zou, Jianke},
  booktitle={2024 6th International Conference on Data-driven Optimization of Complex Systems (DOCS)}, 
  title={Research on Travel Route Planning Optimization based on Large Language Model}, 
  year={2024},
  volume={},
  number={},
  pages={352-357},
  doi={10.1109/DOCS63458.2024.10704489}}

@article{marcelyn2025pathgpt,
  title={PathGPT: Leveraging Large Language Models for Personalized Route Generation},
  author={Marcelyn, Steeve Cuthbert and Gao, Yucen and Zhang, Yuzhe and Gao, Xiaofeng and Chen, Guihai},
  journal={arXiv preprint arXiv:2504.05846},
  year={2025}
}

@inproceedings{Giannotti2007TrajectoryPM,
  author = {Fosca Giannotti and Mirco Nanni and Fabio Pinelli and Dino Pedreschi},
  title = {Trajectory pattern mining},
  booktitle = {KDD},
  year = {2007},
  pages = {330-339}
}

@inproceedings{
shi2024graphconstrained,
title={{GRAPH}-{CONSTRAINED} {DIFFUSION} {FOR} {END}-{TO}-{END} {PATH} {PLANNING}},
author={Dingyuan Shi and Yongxin Tong and Zimu Zhou and Ke Xu and Zheng Wang and Jieping Ye},
booktitle={The Twelfth International Conference on Learning Representations},
year={2024},
url={https://openreview.net/forum?id=vuK8MhVtuu}
}

@article{ZHANG2023103176,
title = {Route planning using divide-and-conquer: A GAT enhanced insertion transformer approach},
journal = {Transportation Research Part E: Logistics and Transportation Review},
volume = {176},
pages = {103176},
year = {2023},
issn = {1366-5545},
doi = {https://doi.org/10.1016/j.tre.2023.103176},
url = {https://www.sciencedirect.com/science/article/pii/S1366554523001643},
author = {Pujun Zhang and Shan Liu and Jia Shi and Liying Chen and Shuiping Chen and Jiuchong Gao and Hai Jiang},
}

@inproceedings{
  dihan2025mapeval,
  title={MapEval: A Map-Based Evaluation of Geo-Spatial Reasoning in Foundation Models},
  author={Mahir Labib Dihan and MD Tanvir Hassan and MD TANVIR PARVEZ and Md Hasebul Hasan and Md Almash Alam and Muhammad Aamir Cheema and Mohammed Eunus Ali and Md Rizwan Parvez},
  booktitle={Forty-second International Conference on Machine Learning},
  year={2025},
  url={https://openreview.net/forum?id=hS2Ed5XYRq}
}

@misc{chai2023graphllmboostinggraphreasoning,
      title={GraphLLM: Boosting Graph Reasoning Ability of Large Language Model}, 
      author={Ziwei Chai and Tianjie Zhang and Liang Wu and Kaiqiao Han and Xiaohai Hu and Xuanwen Huang and Yang Yang},
      year={2023},
      eprint={2310.05845},
      archivePrefix={arXiv},
      primaryClass={cs.CL},
      url={https://arxiv.org/abs/2310.05845}, 
}

@misc{hasan2025mapagenthierarchicalagentgeospatial,
      title={MapAgent: A Hierarchical Agent for Geospatial Reasoning with Dynamic Map Tool Integration}, 
      author={Md Hasebul Hasan and Mahir Labib Dihan and Mohammed Eunus Ali and Md Rizwan Parvez},
      year={2025},
      eprint={2509.05933},
      archivePrefix={arXiv},
      primaryClass={cs.AI},
      url={https://arxiv.org/abs/2509.05933}, 
}

@misc{roberts2023gpt4geolanguagemodelsees,
      title={GPT4GEO: How a Language Model Sees the World's Geography}, 
      author={Jonathan Roberts and Timo Lüddecke and Sowmen Das and Kai Han and Samuel Albanie},
      year={2023},
      eprint={2306.00020},
      archivePrefix={arXiv},
      primaryClass={cs.CL},
      url={https://arxiv.org/abs/2306.00020}, 
}

@misc{balsebre2024lamplanguagemodelmap,
      title={LAMP: A Language Model on the Map}, 
      author={Pasquale Balsebre and Weiming Huang and Gao Cong},
      year={2024},
      eprint={2403.09059},
      archivePrefix={arXiv},
      primaryClass={cs.CL},
      url={https://arxiv.org/abs/2403.09059}, 
}

@misc{xie2024travelplannerbenchmarkrealworldplanning,
      title={TravelPlanner: A Benchmark for Real-World Planning with Language Agents}, 
      author={Jian Xie and Kai Zhang and Jiangjie Chen and Tinghui Zhu and Renze Lou and Yuandong Tian and Yanghua Xiao and Yu Su},
      year={2024},
      eprint={2402.01622},
      archivePrefix={arXiv},
      primaryClass={cs.CL},
      url={https://arxiv.org/abs/2402.01622}, 
}

@misc{latif20243pllmprobabilisticpathplanning,
      title={3P-LLM: Probabilistic Path Planning using Large Language Model for Autonomous Robot Navigation}, 
      author={Ehsan Latif},
      year={2024},
      eprint={2403.18778},
      archivePrefix={arXiv},
      primaryClass={cs.RO},
      url={https://arxiv.org/abs/2403.18778}, 
}

@misc{xiao2025llmadvisorllmbenchmarkcostefficient,
      title={LLM-Advisor: An LLM Benchmark for Cost-efficient Path Planning across Multiple Terrains}, 
      author={Ling Xiao and Toshihiko Yamasaki},
      year={2025},
      eprint={2503.01236},
      archivePrefix={arXiv},
      primaryClass={cs.RO},
      url={https://arxiv.org/abs/2503.01236}, 
}

@inproceedings{deng2025can,
  title={Can llm be a good path planner based on prompt engineering? mitigating the hallucination for path planning},
  author={Deng, Hourui and Zhang, Hongjie and Ou, Jie and Feng, Chaosheng},
  booktitle={International Conference on Intelligent Computing},
  pages={3--15},
  year={2025},
  organization={Springer}
}

@misc{doma2024llmenhancedpathplanningsafe,
      title={LLM-Enhanced Path Planning: Safe and Efficient Autonomous Navigation with Instructional Inputs}, 
      author={Pranav Doma and Aliasghar Arab and Xuesu Xiao},
      year={2024},
      eprint={2412.02655},
      archivePrefix={arXiv},
      primaryClass={cs.RO},
      url={https://arxiv.org/abs/2412.02655}, 
}

@misc{yuan2025llmapllmassistedmultiobjectiveroute,
      title={LLMAP: LLM-Assisted Multi-Objective Route Planning with User Preferences}, 
      author={Liangqi Yuan and Dong-Jun Han and Christopher G. Brinton and Sabine Brunswicker},
      year={2025},
      eprint={2509.12273},
      archivePrefix={arXiv},
      primaryClass={cs.AI},
      url={https://arxiv.org/abs/2509.12273}, 
}

@inproceedings{hong2023metagpt,
  title={MetaGPT: Meta programming for a multi-agent collaborative framework},
  author={Hong, Sirui and Zhuge, Mingchen and Chen, Jonathan and Zheng, Xiawu and Cheng, Yuheng and Wang, Jinlin and Zhang, Ceyao and Wang, Zili and Yau, Steven Ka Shing and Lin, Zijuan and others},
  booktitle={The Twelfth International Conference on Learning Representations},
  year={2023}
}

@inproceedings{gu2025structext,
  title={Structext-eval: Evaluating large language model’s reasoning ability in structure-rich text},
  author={Gu, Zhouhong and Ye, Haoning and Chen, Xingzhou and Zhou, Zeyang and Feng, Hongwei and Xiao, Yanghua},
  booktitle={Proceedings of the 63rd Annual Meeting of the Association for Computational Linguistics (Volume 1: Long Papers)},
  pages={223--244},
  year={2025}
}

@inproceedings{martens_complexity_2022,
	address = {New York, NY, USA},
	series = {{PODS} '22},
	title = {The {Complexity} of {Regular} {Trail} and {Simple} {Path} {Queries} on {Undirected} {Graphs}},
	isbn = {978-1-4503-9260-0},
	url = {https://dl.acm.org/doi/10.1145/3517804.3524149},
	doi = {10.1145/3517804.3524149},
	urldate = {2026-04-18},
	booktitle = {Proceedings of the 41st {ACM} {SIGMOD}-{SIGACT}-{SIGAI} {Symposium} on {Principles} of {Database} {Systems}},
	publisher = {Association for Computing Machinery},
	author = {Martens, Wim and Popp, Tina},
	month = jun,
	year = {2022},
	pages = {165--174},
}

@article{casel_fine-grained_2023,
	title = {Fine-{Grained} {Complexity} of {Regular} {Path} {Queries}},
	volume = {Volume 19, Issue 4},
	issn = {1860-5974},
	url = {http://arxiv.org/abs/2101.01945},
	doi = {10.46298/lmcs-19(4:15)2023},
	urldate = {2026-04-18},
	journal = {Logical Methods in Computer Science},
	author = {Casel, Katrin and Schmid, Markus L.},
	month = nov,
	year = {2023},
	note = {arXiv:2101.01945 [cs]},
	pages = {8625},
}


%% file: custom_2.bib
@article{li2023compressing,
  title={Compressing context to enhance inference efficiency of large language models},
  author={Li, Yucheng and Dong, Bo and Lin, Chenghua and Guerin, Frank},
  journal={arXiv preprint arXiv:2310.06201},
  year={2023}
}

@article{islam2024mapcoder,
  title={MapCoder: Multi-Agent Code Generation for Competitive Problem Solving},
  author={Islam, Md Ashraful and Ali, Mohammed Eunus and Parvez, Md Rizwan},
  journal={arXiv preprint arXiv:2405.11403},
  year={2024}
}

@article{yao2023tree,
  title={Tree of thoughts: Deliberate problem solving with large language models},
  author={Yao, Shunyu and Yu, Dian and Zhao, Jeffrey and Shafran, Izhak and Griffiths, Thomas L and Cao, Yuan and Narasimhan, Karthik},
  journal={arXiv preprint arXiv:2305.10601},
  year={2023}
}

@article{COT,
  title={Chain-of-thought prompting elicits reasoning in large language models},
  author={Wei, Jason and Wang, Xuezhi and Schuurmans, Dale and Bosma, Maarten and Xia, Fei and Chi, Ed and Le, Quoc V and Zhou, Denny and others},
  journal={Advances in Neural Information Processing Systems},
  volume={35},
  pages={24824--24837},
  year={2022}
}
